\documentclass{article} 
\usepackage{times}
\usepackage[accepted]{icml2025}

\usepackage{amsmath,amsfonts,bm}

\def\eqref#1{equation~\ref{#1}}
\def\Eqref#1{Equation~\ref{#1}}

\def\1{\bm{1}}

\DeclareMathAlphabet{\mathsfit}{\encodingdefault}{\sfdefault}{m}{sl}
\SetMathAlphabet{\mathsfit}{bold}{\encodingdefault}{\sfdefault}{bx}{n}

\usepackage{hyperref}       
\usepackage{url}            
\usepackage{booktabs}       
\usepackage{amsfonts}       
\usepackage{nicefrac}       
\usepackage{microtype}      
\usepackage{xcolor}         
\usepackage{graphicx}

\usepackage{booktabs,tabularx}
\usepackage{float}
\usepackage{fancyvrb}
\usepackage{amsmath}
\usepackage{amssymb}
\icmltitlerunning{Learning Human-Aligned Representations with Contrastive Learning and Generative Similarity}

\begin{document}

\twocolumn[
\icmltitle{Learning Human-Aligned Representations \\ with Contrastive Learning and Generative Similarity}



\icmlsetsymbol{equal}{*}

\begin{icmlauthorlist}
\icmlauthor{Raja Marjieh}{equal,psych}
\icmlauthor{Sreejan Kumar}{equal,neuro}
\icmlauthor{Declan Campbell}{neuro}
\icmlauthor{Liyi Zhang}{CS}
\icmlauthor{Gianluca Bencomo}{CS}
\icmlauthor{Jake Snell}{CS}
\icmlauthor{Thomas L. Griffiths}{psych,CS}
\end{icmlauthorlist}

\icmlaffiliation{psych}{Department of Psychology, Princeton University}
\icmlaffiliation{neuro}{Princeton Neuroscience Institute}
\icmlaffiliation{CS}{Department of Computer Science, Princeton University}

\icmlcorrespondingauthor{Sreejan Kumar}{sreejank@princeton.edu}
\icmlcorrespondingauthor{Raja Marjieh}{raja.marjieh@princeton.edu}

\icmlkeywords{Representations, Cognitive Science, Human Alignment, Bayesian Inference}

\vskip 0.3in
]

\newcommand{\fix}{\marginpar{FIX}}
\newcommand{\new}{\marginpar{NEW}}

\printAffiliationsAndNotice{\icmlEqualContribution} 

\begin{abstract}
    Humans rely on effective representations to learn from few examples and abstract useful information from sensory data. Inducing such representations in machine learning models has been shown to improve their performance on various benchmarks such as few-shot learning and robustness. However, finding effective training procedures to achieve that goal can be challenging as psychologically rich training data such as human similarity judgments are expensive to scale, and Bayesian models of human inductive biases are often intractable for complex, realistic domains. Here, we address this challenge by leveraging a Bayesian notion of generative similarity whereby two data points are considered similar if they are likely to have been sampled from the same distribution. This measure can be applied to complex generative processes, including probabilistic programs. We incorporate generative similarity into a contrastive learning objective to enable learning of embeddings that express human cognitive representations. We demonstrate the utility of our approach by showing that it can be used to capture human-like representations of shape regularity, abstract Euclidean geometric concepts, and semantic hierarchies for natural images.
\end{abstract}

\section{Introduction}
Human intelligence is characterized by representations that enable humans to form meaningful generalizations \citep{tenenbaum2011grow}, learn from few examples \citep{lake2015human}, and abstract useful information from sensory data \citep{gershman2017blessing}. In cognitive science, probabilistic models based on Bayesian inference are used to explain how humans make meaningful generalizations from small amounts of data \citep{tenenbaum2011grow,griffiths2010probabilistic,piantadosi2016logical,goodman2011learning}. In machine learning, many recent efforts have gone towards aligning neural network models to human behavior by incorporating psychologically-rich human data in the training objectives, such as reward functions \cite{ouyang2022training,christiano2017deep}, categorization uncertainty \citep{collins2023human,peterson2019human}, language descriptions \citep{radford2021learning, kumar2022using,marjieh2022words}, soft labels \citep{sucholutsky2021soft}, or similarity judgements \citep{muttenthaler2024aligning,muttenthaler2024improving}. 

\begin{figure}[t]
  \centering  \includegraphics[width=\linewidth]{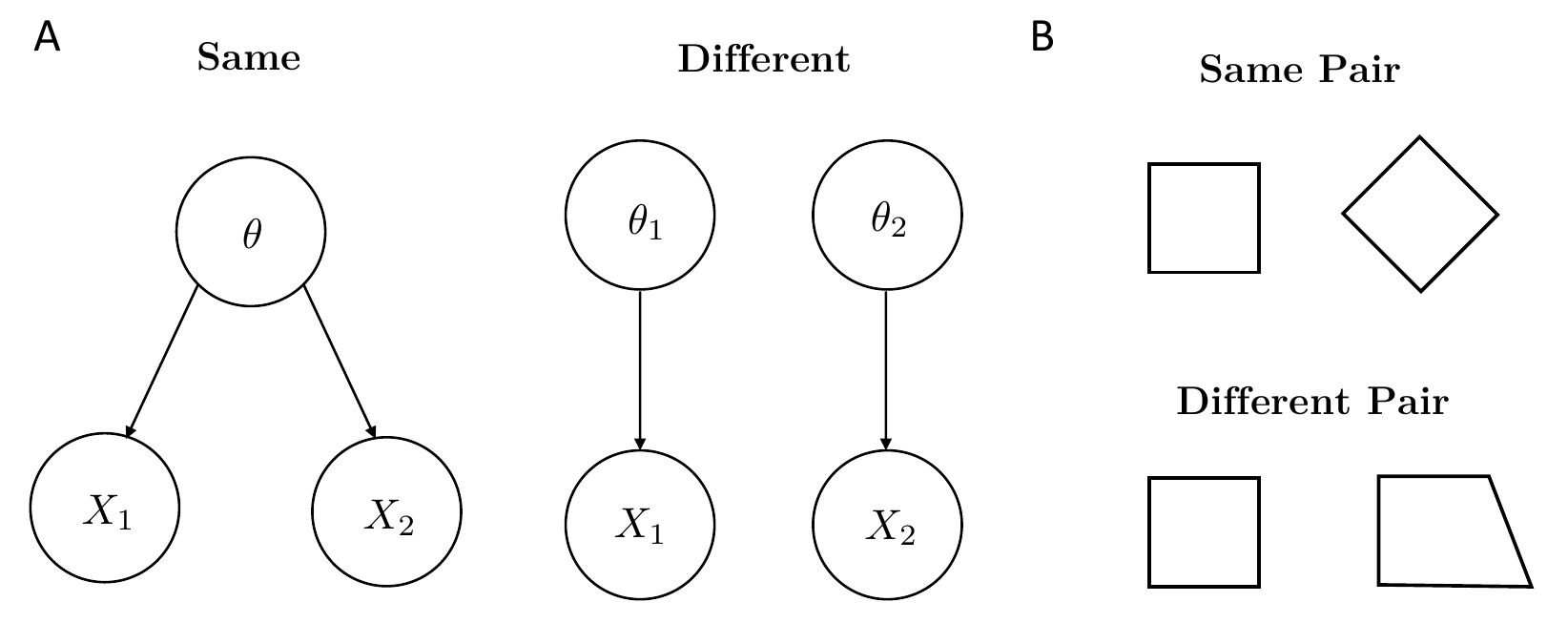}
  \caption{\textbf{Schematic representation of generative similarity.} \textbf{A.} Graphical models for the same and different data generation hypotheses. \textbf{B.} Example same and different quadrilateral shape pairs.}
  \label{fig:schematic}
\end{figure}

Both approaches have limitations. Though Bayesian models provide an effective description of human behavior, they are often computationally intractable due to expensive posterior computations that require summing over large hypothesis spaces. Incorporating human judgments in model training objectives, such as human similarity judgments, may be intuitive for alignment \citep{esling2018generative,jha2023extracting,muttenthaler2024improving}, but collecting them at the scale necessary to train large machine learning models can become challenging as the number of required judgments grows quadratically in the number of stimuli (though see \citet{marjieh2022words} for proxies).

Here, we introduce a new technique for learning human-aligned representations that combines the strengths of these two approaches by integrating Bayesian models of similarity with contrastive learning \citep{chen2020simple}. Contrastive learning uses the designation of datapoints as being the ``same'' or ``different'' to learn a representation where the ``same'' datapoints are encouraged to be closer together and ``different'' datapoints further apart. This approach provides a way to go from a similarity measure to a representation. We define a principled notion of similarity based on Bayesian inference (initially proposed in \citealp{kemp2005generative}) and show that it can be implemented within contrastive learning. Specifically, given a set of samples and a hierarchical generative model from which data distributions are first sampled and then individual samples are drawn, we define the \emph{generative similarity} between a pair of samples to be their probability of having been sampled from the same distribution relative to that of them being sampled from two independently-drawn distributions (Figure \ref{fig:schematic}). Combining generative similarity using models that reflect human behavior with contrastive learning can provide a general procedure for aligning human and machine representations.

To demonstrate the utility of our approach, we apply it to three domains where humans exhibit complex behavior. 
First, we build off work in cognitive science quantifying shape regularity biases \cite{sable2021sensitivity} in human vs. animal intelligence and use generative similarity to instill human-like shape regularity into vision model representations. Second, we use generative similarity over probabilistic programs to produce human-like few-shot learning of abstract geometric concepts. Third, we apply generative similarity to the hierarchically organized categories of ImageNet \citep{deng2009imagenet} to demonstrate learning of hierarchical image categories and capturing of human behavior when reasoning about different abstraction levels. Viewed together, these results highlight a path towards alignment of human and machine intelligence by using Bayesian models of cognition to learn human-aligned representations via a scalable contrastive learning framework.

\section{Combining Generative Similarity with Contrastive Learning}\label{theory}
Given a set of samples $D$ and an associated generative model of the data $p(D)=\int p(D|\theta)p(\theta)d\theta$ where $p(\theta)$ is some prior over distribution parameters (e.g., a beta prior) and $p(D|\theta)$ is an associated likelihood function (e.g., a Bernoulli distribution), we define the \emph{generative similarity} between a pair of samples $s_\text{gen}(x_1,x_2)$, to be the Bayesian odds ratio for the probability that they were sampled from the same distribution to that of them being sampled from two independent (or ``different'') distributions
\begin{align}\label{eq:gen_sim}
&s_\text{gen}(x_1,x_2)=\frac{p(\text{same}|x_{1},x_{2})}{p(\text{different}|x_{1},x_{2})} \\
&\quad\,\,\,\,=\frac{\int p(x_{1}|\theta)p(x_{2}|\theta)p(\theta)d\theta}{\int p(x_{1}|\theta_{1})p(x_{2}|\theta_{2})p(\theta_{1})p(\theta_{2})d\theta_{1}d\theta_{2}} \nonumber
\end{align}
where we assume that \textit{a priori} $p(\text{same})=p(\text{different})$ (i.e., the prior over the two hypotheses is uniform). The same and different data generation hypotheses are shown in Figure~\ref{fig:schematic}A along with example same and different pairs in Figure~\ref{fig:schematic}B in the case of a generative process of quadrilateral shapes, with the same pair corresponding to two squares, and the different pair corresponding to a square and a trapezoid. This definition builds on existing Bayesian models of similarity in cognitive science \citep{shepard1987toward,tenenbaum_griffiths_2001}, and in particular \citet{kemp2005generative}.

Given the definition of generative similarity, we next distinguish between two scenarios. When $s_{\text{gen}}$ can be directly computed, then its incorporation in a contrastive loss function is straightforward: given a parametric neural encoder $\phi_\varphi(x)$ and a prescription for deriving similarities from these embeddings, e.g. $s_\text{emb}=s_0e^{-d}$ where $d$ is a distance measure and $s_0$ is a constant, we can then directly optimize the embedding parameters such that the difference between the generative similarity and the corresponding embedding similarity is minimized, e.g.,
\begin{equation}\label{eq:loss_1}
    \varphi^\ast=\arg\min_\varphi \mathbb{E}_{p}(s_\text{emb}(\phi_\varphi(X_1),\phi_\varphi(X_2))-s_\text{gen}(X_1,X_2))^2.
\end{equation} 
In cases where $s_\text{gen}$ cannot be computed, we can implicitly incorporate it in training by applying Monte Carlo sampling from the generative model into individual triplet loss functions (see Appendix \ref{app:triplet} for more details and Appendix \ref{app:gaussian_triplet_exp} for example simulations in an analytically tractable use case). For some domains it is also possible to use available proxies (kernels) for distributional similarity as we will show later.



\subsection{Related Work}

We next discuss how our approach connects to existing methods. First, observe that by carrying out the integration in \Eqref{eq:gen_sim} we have $s_{\text{gen}}(x_1,x_2)=p(x_1,x_2)/p(x_1)p(x_2)$ where $p(x_1,x_2)=\int p(x_{1}|\theta)p(x_{2}|\theta)p(\theta)d\theta$ and $p(x_{1,2})$ are its marginals. This means that (up to a logarithm) generative similarity is a specific type of point-wise mutual information \citep{church1990word}. While point-wise mutual information (MI) is broadly concerned with sample independence, generative similarity goes further by tying this independence to a hierarchical generative structure that encapsulates an inductive bias associated with the data. This is in line with work suggesting that the success of MI-based contrastive learning approaches hinges significantly on the specific inductive biases incorporated in the process \citep{tschannen2019mutual}. Notable contrastive methods with losses based on MI maximization include SimCLR \citep{chen2020simple} and InfoNCE \citep{oord2018representation} more broadly. SimCLR is a training framework whereby a representation is learned by applying various augmentations to the training data and then using those to construct positive and negative pairs that are then incorporated in a contrastive (e.g. triplet) loss. Crucially, positive and negative pairs in SimCLR (and InfoNCE) are usually constructed in a domain-agnostic manner by applying augmentations such as rotations and rescalings to the data. While this is a good default strategy, when there is prior expectation about the structure of the underlying generative process, it glosses over finer structure present in the data, such as within- and across-category relations. Supervised contrastive learning \cite{khosla2020supervised} goes one step in this direction by defining similarity with respect to meaningful semantic labels. Our approach expands this idea by moving from similarity over labels to similarity over distributions, which can capture more nuanced, complex, and graded relationships among datapoints that go beyond those of simple independent, discrete categories. The experiments we present show how training models with such relationships in mind can lead to greater alignment with human behavior. 


Second, it is possible to show under suitable regularization (to ensure that $s_{\text{gen}}$ can be treated as a distribution) that generative similarity (\Eqref{eq:gen_sim}) can be derived as the minimizer of the functional $D_{KL}[s||p_{\text{same}}]-D_{KL}[s||p_{\text{diff}}]$ (see Appendices~~\ref{app:opt-gen-sim}-\ref{app:lin}) which is reminiscent of contrastive divergence learning, a method based on a contrastive objective between two Kullback-Leibler (KL) divergences \citep{carreira2005contrastive}. However, while the above objective is also a contrastive difference between divergences, it is different from that found in CDL in which the goal is to reduce the difference between the divergences whereas in our case it is to increase their (negative) contrast.

Finally, there are other contrastive frameworks that incorporate a Bayesian component. Here it is worth noting two lines of work: (i) Contrastive kernel methods such as the mutual information kernel (MIK) \citep{seeger2001covariance} and the positive pair kernel (PPK) \citep{johnson2022contrastive}. Both MIK and PPK involve a kernel object that is analogous to \Eqref{eq:gen_sim} as the basis of their approach. However, they both differ from our current framework, the former being largely a support vector machine, and the latter considering generative processes in the limited sense of domain-agnostic augmentations (as with SimCLR) and without explicitly incorporating the kernel in a contrastive loss (as with Equation \ref{eq:loss_1}). It is also unclear from these works how such contrastive kernels can be used to align human and machine representations. (ii) Other hybrid Bayesian contrastive learning methods (e.g., \citealp{liu2023bayesian}) devise Bayesian computations to mine hard negative examples to improve contrastive training. However, these are not associated with a generative model as ours is.


\section{Experiments}

\subsection{Instilling Representations of Shape Regularity }

\begin{figure*}[t]
  \centering  \includegraphics[width=\linewidth]{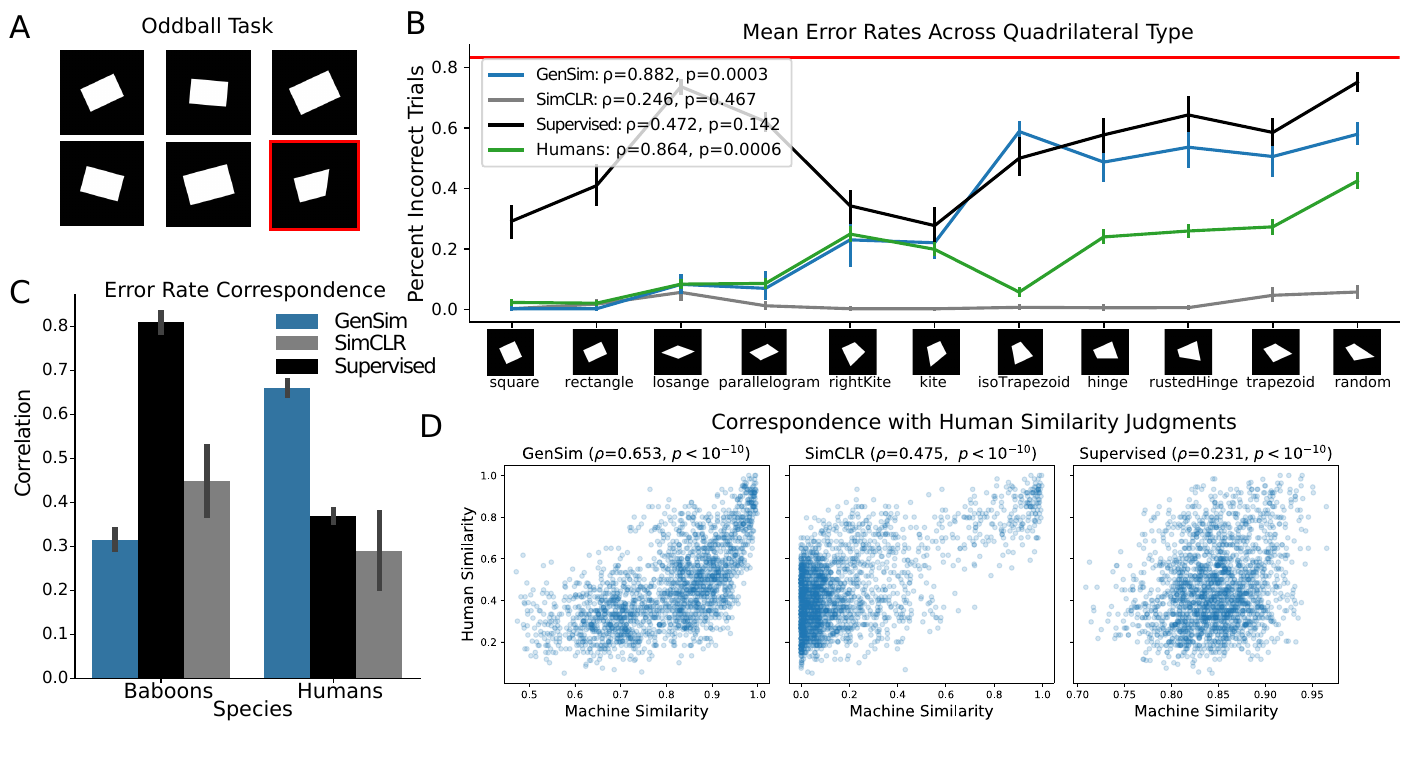}
  \caption{\textbf{Generative similarity can instill human shape regularity biases.} \textbf{A.} The oddball task of \citet{sable2021sensitivity} used six quadrilateral stimulus images, in which five images were of the same reference shape (differing in scale and rotation) and one was an oddball (highlighted in red) that diverged from the reference shape’s geometric properties. In this example, the reference shape is a rectangle; note that the oddball does not have four right angles like the rectangles. \textbf{B.} Mean error rates per quadrilateral type on the oddball task, ordered from most regular to least regular shapes. \citet{sable2021sensitivity} found that human error rates systematically increase as the regularity of the reference quadrilaterals decrease, whereas those of non-human primates do not. We use this same evaluation for a CNN model with different finetuning objectives, reporting the Spearman rank correlation between model performance and number of geometric regularities across quadrilateral type (see Table 1 in Appendix~\ref{app:quad-detail} for these values). Only the model with the generative similarity (GenSim) finetuning objective exhibited a significant human geometric regularity effect. Error bars denote standard errors over $10$ training runs, red line shows chance-level error rate. \textbf{C.} Correlation between model error rates and human or monkey error rates. Supervised is the most monkey-like whereas GenSim is the most human-like. \textbf{D.} We show the Spearman rank correlation of pairwise human similarity judgements of quadrilaterals to those of the models. }
  \label{fig:quads}
\end{figure*}

{\bf Background:}
Psychological research suggests that the human species is uniquely sensitive to abstract shape regularity \citep{henshilwood2011100,saito2014origin}.  \citet{sable2021sensitivity} compared diverse human groups (varying in education, cultural background, and age) to non-human primates on a simple oddball discrimination task. Participants were shown a set of five reference shapes and one ``oddball'' shape and were prompted to identify the oddball (Figure~\ref{fig:quads}A). The reference shapes were generated using basic geometric regularities: parallel lines, equal sides, equal angles, and right angles, which can be specified by a binary vector corresponding to the presence or absence of these specific geometric features. There were 11 types of quadrilateral reference shapes with varying geometric regularity, from squares (most regular) to random quadrilaterals containing no parallel lines, right angles, or equal angles/sides (least regular; Figure~\ref{fig:quads}B). In each trial, five different versions of the same reference shape (e.g., a square) were shown in different sizes and orientations. The oddball shape was a modified version of the reference shape, in which the lower right vertex was moved such that it violated the regularity of the original reference shape (e.g., moving the lower right vertex of a trapezoid such that it no longer has parallel sides). Figure~\ref{fig:quads}A shows an example trial.

\citet{sable2021sensitivity} found that humans were sensitive to these geometric regularities (right angles, parallelism, symmetry, etc.) whereas non-human primates were not. Specifically, they found that human performance was best on the oddball task for the most regular shapes, and systematically decreased as shapes became more irregular (Figure~\ref{fig:quads}B). Conversely, non-human primates performed well above chance, but they performed worse than humans overall and, critically, exhibited no influence of geometric regularity. Additionally, they tested a pretrained convolutional neural network (CNN) model, CorNet \citep{kubilius2019brain}, on the task. CorNet (Core Object Recognition Network) is a convolutional neural network model with an architecture that models the primate ventral visual stream. It is pretrained on a standard supervised object recognition objective on ImageNet and is one of the top-scoring models of ``brain-score'', a benchmark for testing models of the visual system using both behavioral and neural data  \citep{schrimpf2018brain}. Like the monkeys, a pretrained CorNet exhibited no systematic relationship with the level of geometric regularity.

{\bf Generative Similarity:}
Shape regularity serves as an ideal case study for our framework because i) it admits a generative similarity measure that can be computed in closed form, and ii) we can use it to test whether our contrastive training framework can induce the human shape regularity effect observed by \citet{sable2021sensitivity} in a neural network. Recall that the shape categories of \citet{sable2021sensitivity} can be specified by binary feature vectors corresponding to the presence or absence of abstract geometric features (equal angles, equal sides, parallel lines, and right angles of the quadrilateral) from which individual examples (or exemplars) can be sampled. Formally, we can define a natural generative process for such shapes as follows: given a set
of binary geometric feature variables $\{F_{1},\dots,F_{n}\}\in\{0,1\}^{n}$, we define
a hierarchical distribution over shapes by first sampling Bernoulli
parameters $\theta_{i}$ for each feature variable $F_{i}$ from a prior
$\text{Beta}(\alpha,\beta)$, then sampling
feature values $f=(f_{1},\dots,f_{n})$ from the resulting Bernoulli
distributions $\text{Bern}(\theta_{i})$, and then uniformly sampling
a shape $\sigma(f)$ from a (possibly large) list of available exemplars $\mathcal{S}(f)=\{\sigma_{1}(f),\dots,\sigma_{M}(f)\}$
that are consistent with the sampled feature vector $f$ (the set could also be empty if the geometric features are not realizable due to geometric constraints). In other words, the
generative process is defined as
\begin{equation}
\begin{cases}
\theta_{i}\sim\text{Beta}(\alpha,\beta), & \text{sample Bernoulli parameters}\\
f_{i}\sim\text{Bern}(\theta_{i}), & \text{sample discrete features}\\
\sigma(f)\sim\text{Uniform}(\mathcal{S}(f)), & \text{sample shape exemplar}
\end{cases}\label{eq:shape-process}
\end{equation}
This process covers both soft and definite categories, and our current setting corresponds to the special limit $\alpha=\beta\rightarrow0$, in which case the Beta prior over Bernoulli parameters becomes concentrated around $0$ and $1$ so that the process becomes that of choosing a category specified by a set of geometric attributes and then sampling a corresponding exemplar. In Appendix \ref{app:shape-deriv}, we use the conjugacy relations between the Beta and Bernoulli distributions to derive the generative similarity associated with the process in \Eqref{eq:shape-process}, and we show that in our limit of interest ($\alpha=\beta\rightarrow0$) the corresponding generative similarity measure between shapes $\sigma_1,\sigma_2$ with feature vectors $f_i^{(1)},f_i^{(2)}$ is given by the formula
    $\log s(\sigma_{1}(f^{(1)}),\sigma_{2}(f^{(2)}))\propto-\sum_{i}(f_{i}^{(1)}-f_{i}^{(2)})^{2}$.

We used this generative similarity measure to finetune CorNet, the same model \citet{sable2021sensitivity} used in their experiments, to see whether our measure would induce the human geometric regularity bias (Figure~\ref{fig:quads}). Specifically, given a random pair of quadrilateral stimuli, we computed the above quantity (i.e. the Euclidean distance) between their respective binary geometric feature vectors (presence and absence of equal sides, equal angles, and right angles) and finetuned the pretrained CorNet model on a contrastive learning objective using these distances. This pushed quadrilaterals with similar geometric features together and pulled those with different geometric features apart in the model's representation (additional details regarding training are provided in Appendix \ref{app:quad-detail}). Like \citet{sable2021sensitivity}, to test the model on the oddball task, we extract the embeddings for all $6$ choice images and choose the oddball as the one that is furthest (i.e. Euclidean distance) from the mean embedding.

{\bf Results:}
The geometric regularity effect observed for humans in  \citet{sable2021sensitivity} was an inverse relationship between geometric regularity and error rate. For example, humans performed best on the most regular shapes, such as squares and rectangles and worst on the least regular shapes, such as trapezoids. This regularity effect was absent in the monkey and pretrained CorNet error rates. In Figure~\ref{fig:quads}B, we show error rates as a function of geometric regularity on a CorNet model finetuned on generative similarity (GenSim CorNet; blue line). We compare this model to two baselines. First, we show the performance of a CorNet model finetuned on a supervised classification objective on the quadrilateral stimuli, where the model must classify which of the $11$ categories a quadrilateral belongs to. This is the same objective \citet{sable2021sensitivity} used to finetuned CorNet, and we replicate their results here as a baseline for our proposed method. Second, we use a standard contrastive learning objective, SimCLR \citep{chen2020simple}, which produces augmented versions of an image with the goal of making representations of an image and its augmented version as similar as possible. Following the original SimCLR work, the augmentations we used were random cropping, rotations, and gaussian blurring. All models were trained for $13$ epochs (the same number of epochs used by \citealp{sable2021sensitivity}). 

Like humans, the model error rates for the GenSim CorNet model significantly increase as the shapes become more irregular (Spearman's $\rho(9)=0.88,p=0.0003$). This is not the case for the model finetuned with supervised classification (Spearman's $\rho(9)=0.472,p=0.142$). We also correlated the error rates of the finetuned models with those of humans and monkeys (Figure~\ref{fig:quads}C) and see a double dissociation between the two models. Specifically, the generative similarity-trained CorNet model's error rates match human error rates significantly more than monkey error rates, $t(18)=12.45,p<0.0001$, whereas those of the baseline supervised CorNet model match monkey error rates significantly more than human error rates (Figure~\ref{fig:quads}C), $t(18)=17.43,p<0.0001$. This double dissociation is consistent across different subjects (see Supplementary Figure~\ref{fig:consistency}). To further confirm the superior alignment of the GenSim model to human representations, we collected a dataset of pairwise human similarity judgments of the reference quadrilaterals and correlated these judgments to those of the models (i.e., embedding cosine similarty; Figure~\ref{fig:quads}D), where we found that GenSim has the highest correspondance to human similarity judgments (see Appendix \ref{human-exp} for details regarding data collection). 

Finetuning on the SimCLR objective does not result in the human geometric regularity effect ($\rho=0.246,p=0.467$). The performance of SimCLR is much higher than that of other models (Figure~\ref{fig:quads}B) and both humans and baboons. To understand why, recall that the reference image choices in the Oddball task are different scales/rotations of the reference image. Because the SimCLR objective applies similar image augmentations to minimize the distance of an image's embedding with its augmented counterpart, the overlap between the training paradigm and the Oddball task allows the network to overfit to it. Crucially, however, this network is not \textit{human-like} because it lacks the human shape regularity effect that the GenSim model possesses (Figure~\ref{fig:quads}B-C).

\begin{figure*}[t]
  \centering  \includegraphics[width=\linewidth]{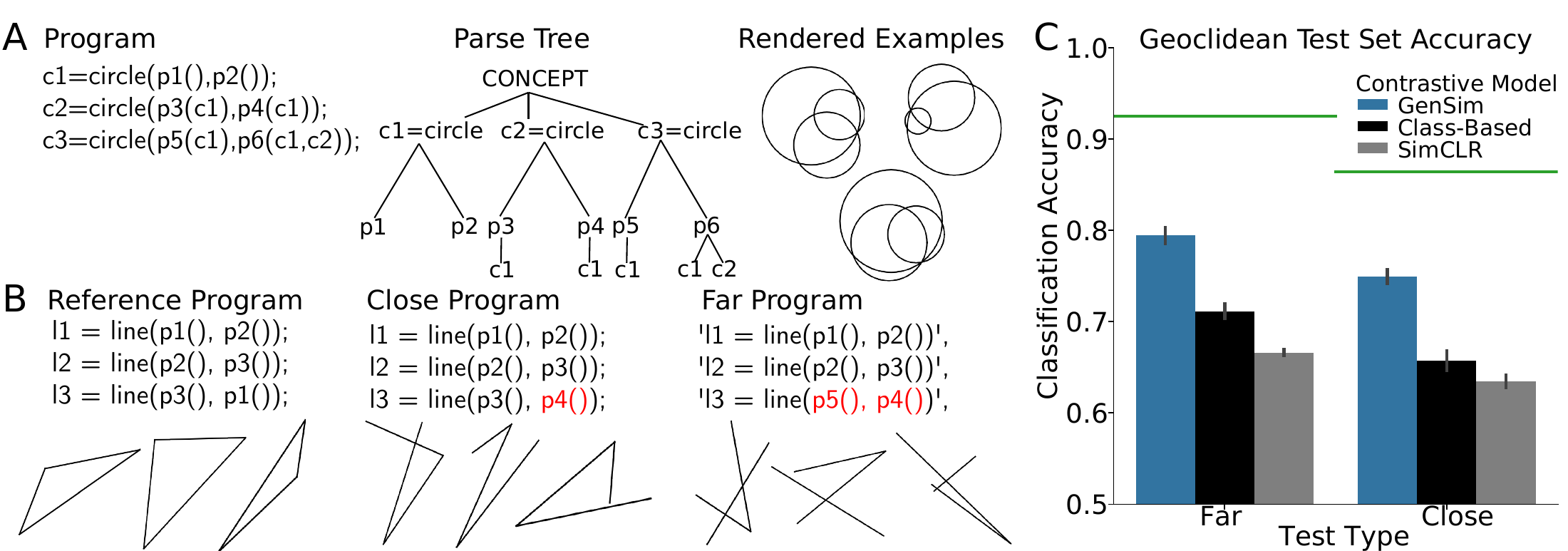}
  \caption{\textbf{Generative similarity over probabilistic programs helps contrastive learning models better represent geometric concepts.} \textbf{A.} Example program, program parse tree, and rendered generations from the Geoclidean DSL. \textbf{B.} In the Geoclidean classification task, participants or models are given examples from a reference program. They have to successfully distinguish between examples from the reference program and two types of negative examples - ones from a ``close'' program and ones from a ``far'' program.  \textbf{C.} Performance of contrastive embeddings in correctly classifying reference program images. Green lines indicate mean human accuracy and errorbars denote standard error over training runs. }
  \label{fig:programs}
\end{figure*}
\subsection{Learning Abstract Geometric Concepts through Probabilistic Programs}
{\bf Background:}
Our next test case builds on recent work that shows the limitations of current vision models in understanding abstract concepts of geometric stimuli, a capability that humans possess \citep{hsu2022geoclidean}. \citet{hsu2022geoclidean} proposed a benchmark based on a probabilistic programming language that synthesizes stimuli that capture intuitive geometric concepts. The Geoclidean Domain-Specific Language (DSL) is directly inspired by the work of Euclid, who formalized geometric axioms using a compass and a straight edge. The DSL contains three simple primitives reflecting the principles of Euclidean geometry -- a point with an $x,y$ location, a line, which is parameterized by two points, and a circle, which is parameterized by a center point and an edge point. Lines and circles are parametrized by points while points can be bound to previously built objects. This DSL can define a distribution of probabilistic programs that can then be rendered into images (see Figure~\ref{fig:programs}A). Each program can be thought of as the specification for a set of geometric objects that have complex relations among them. 

\citet{hsu2022geoclidean} define a set of $37$ programs within this DSL that resemble concepts defined in Euclid's \textit{Elements} \cite{simson1838elements}. They then design a classification task where participants or models are given examples from a reference program and have to use these examples to correctly classify other images from the reference programs (positive examples) and images from different programs (negative examples). There were two types of programs that rendered the negative examples (Figure~\ref{fig:programs}B), ``close'', which had minimal changes with respect to the reference program, and ``far'', which had a greater number of changes to the reference program. This task is designed to probe the intuitive understanding of geometry, as evidenced by the ability to generalize from a limited set of examples to new instances. \citet{hsu2022geoclidean} demonstrated that humans show a robust capacity for such generalization across 37 different concepts, suggesting a natural sensitivity to the hierarchical structure and part relations inherent in Euclidean geometry. Conversely, they found that pretrained vision models struggle with few-shot generalization in this context.

{\bf Generative Similarity:} The syntactic structure of the Geoclidean programs sufficiently describes the relations between the primitive objects (point, circle, and line) necessary to correctly categorize images as belonging to a particular program. Given this, we formulated a Context-Free Grammar (CFG) from the Geoclidean DSL that can parse the Geoclidean programs into parse trees (see Appendix~\ref{app:geoclidean_cfg} for the full CFG grammar). If we interpret the Geoclidean programs as samples from this CFG, their generative similarity will be based on sharing the same production rules or, equivalently, shared substructures in their parse trees. As an implementation of this measure, we used the Fast Tree Kernel (FTK) by \citet{moschitti2006making}, which recursively computes the number of shared subtrees in two parse trees to determine their similarity. The loss function for generative similarity is, therefore, the mean squared error between the cosine similarity between two rendered images' embeddings and the tree kernel similarity between the images' respective generative programs. 

We used the $37$ geometric concepts presented in the original Geoclidean work for our experiments (see Appendix~\ref{app:concepts} for the full list and their respective programs). We randomly held-out $7$ concepts for testing our models and generated a dataset of $100,000$ images that were random samples from the $30$ training concepts (around $3000$ images per concept). We used a pretrained ResNet-50 architecture for our experiments \cite{he2015deep}, a model shown in the original Geoclidean work to do poorly at the benchmark. We had two baseline models, which used the same architecture and training data. The first baseline model was the standard contrastive learning objective used in the SimCLR paper \citep{chen2020simple}. The SimCLR objective produces augmented versions of an image (e.g. random cropping, rotations, gaussian blurring, etc.) with the goal of making representations of an image and its augmented version as similar as possible and representations of different images as dissimilar as possible. This objective is equivalent to a generative similarity measure for a default generative process devoid of domain-specific knowledge in which a base image is sampled and then a random augmentation is applied. The second baseline is a model that uses class information during training. In particular, the loss is a triplet-based loss where the anchor image and positive image are samples from the same program and the negative image is a sample from a different program. This loss is equivalent to a generative similarity measure where the underlying generative model is a discrete set of $30$ categories corresponding to each of the training programs and images sampled from those programs are members of that category. It is also equivalent to using supervised contrastive learning \cite{khosla2020supervised} where the $30$ training programs are the human-labeled classes used for supervision. By using these baselines, we show the value of providing richer generative models in producing networks aligned towards human-like capabilities such as few-shot generalization of geometric concepts. See Appendix \ref{app:draw-detail} for more details.

{\bf Results: } We evaluated our models using the exact same evaluation approach used by \citet{hsu2022geoclidean} on the held-out test programs (Figure~\ref{fig:programs}C). In particular, we computed the mean embedding across $5$ reference images from a test program (``the prototype'') and measured the normalized $\ell_{1}$ distance between the test image's embedding and the program prototype. A threshold was fit to each test concept and model to maximize classification accuracy (\citet{hsu2022geoclidean} did this to bias accuracy in favor of the models over humans, and we replicate this process across all of our models to ensure fair comparison across models). The GenSim model, though still below human accuracy, performs the best at classification, followed by the class-based and SimCLR models. To our knowledge, this is the current state-of-the-art accuracy on the Geoclidean benchmark, although it is important to note that \citet{hsu2022geoclidean} did not finetune their models on the task.

\subsection{Capturing Hierarchical Categories in Natural Images with Generative Similarity}\label{im-hierarchy}

\begin{figure*}[h]
  \centering  
  \includegraphics[width=\linewidth]{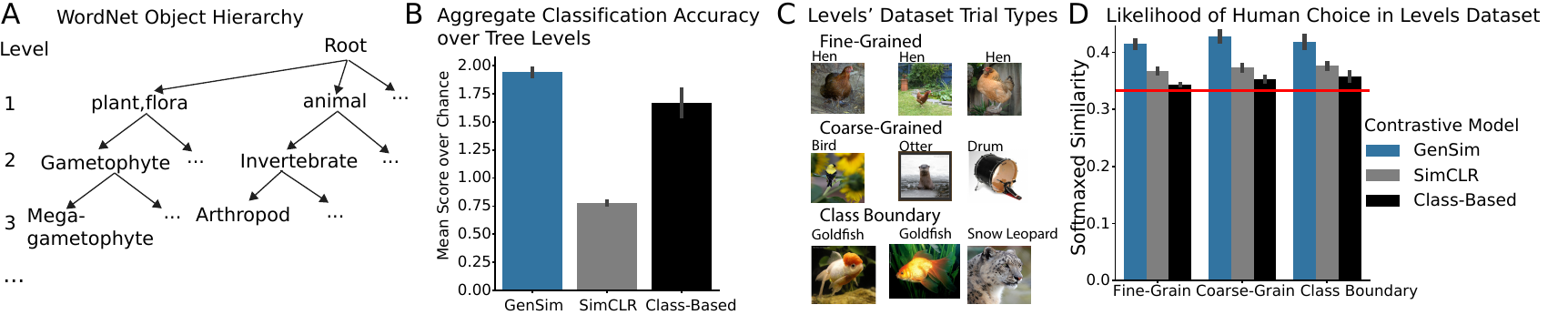}
  \caption{\textbf{Generative Similarity Helps Encode Natural Image Hierarchies} \textbf{A.} WordNet \cite{church1990word} gives a hierarchy that spans multiple semantic levels. The object categories in the classic ImageNet dataset \cite{deng2009imagenet} are sampled from different places on the WordNet tree. \textbf{B.} Mean classification accuracy over chance across tree levels. Error bars denote standard errors over model training runs. See Supplementary Figure~\ref{fig:tree_level_acc} for accuracies on individual tree levels. \textbf{C.} The Levels Dataset of \citet{muttenthaler2024improving} collected human odd-one-out judgments at $3$ different abstraction levels where humans are given three images and choose what they believe is the odd-one-out. \textbf{D.} Mean model likelihood across human participants calculated using each model's pairwise embeddings' similarities. We found that the GenSim model best accounts for human behavior. Error bars are standard errors across human subjects. }
  \label{fig:imagenet}
\end{figure*}

{\bf Background:} An important facet of human intelligence is the ability to formulate abstract categories that allow us to generalize across individual stimuli. For example, when encountering a dog for the first time, we may come up with the abstract category of “animal” and generalize this concept to different instances, such as a cat. Humans naturally organize such abstract categories into complex hierarchies  (e.g., living organism $\rightarrow$ animal $\rightarrow$ mammal $\rightarrow$ dog; \citealp{tenenbaum2011grow,murphy2004big}, Figure~\ref{fig:imagenet}A). This hierarchical structure is reflected in many big machine learning datasets, including ImageNet  classes \citep{deng2009imagenet}, though many common models trained on this dataset assume a flat distribution over independent categories without necessarily taking into account their relations with one another in the hierarchy. 

{\bf Generative Similarity:} WordNet \cite{church1990word} is a large lexical database of English words that hierarchically organizes words and concepts. The ImageNet classes are based on words from WordNet. 
We derived a measure of generative similarity using the WordNet hierarchy as a part of a generative model (Figure~\ref{fig:imagenet}A). In this generative model, we randomly sample a category by traversing the tree, where each path has equal probability, and then randomly sample an example from that category. A single image can be described by multiple categories differing on the tree level (e.g. a spider image is also an arthropod as well as an invertebrate and an animal). Let $i$ and $j$ be the lowest possible categories for images $x_i$ and $x_j$ (i.e. the leaf nodes on the WordNet tree or, equivalently, the classic category used in the ImageNet dataset). Our measure of generative similarity between the two images is as follows (see Appendix~\ref{app:hierarchy_formula_deriv} for a formal derivation)
\begin{equation}\label{imagenetformula}
    s(x_{i},x_{j})=\frac{1}{K}[\frac{1}{p_{1}}+\frac{1}{p_{1}p_{2}}+\dots+\frac{1}{p_{1}p_{2}\dots p_{K}}]
\end{equation}
where $K$ is the depth of the highest common ancestor between $x_{i}$ and $x_{j}$ (``separation depth'') and $p_{1\dots K}$ are the probabilities of choosing each edge along the path (which we take to be uniform per level). The loss function based on generative similarity involves sampling two images, extracting the cosine similarity between their embeddings, and computing the mean squared error between the cosine similarity and $\log[s(x_{i},x_{j})]$. To show that generative similarity with contrastive learning can enable learning of embeddings that encode hierarchal image categories, we trained a ResNet-50 model on the standard ImageNet-1k dataset using this loss function. Following the previous experiment, we had two baselines using the same architecture and training data -- one where we employ a triplet-based loss based on the image category (one anchor image, one positive image from the same class, and one negative image from a different class) and a SimCLR loss. Within our generative similarity framework, the former corresponds to a generative process that involves an independent set of $1000$ categories for which images can be a member and the latter corresponds to a domain-agnostic process that involves sampling a random image and applying an augmentation. Additional details are provided in Appendix \ref{app:imagenet-detail}.  

{\bf Results:}
To quantify how much the embeddings have encoded hierarchical categories, we used the learned models' embeddings to train a Linear SVM classifier to classify categories from the embeddings. We did this for every level of the WordNet tree (see left side of Figure~\ref{fig:imagenet}A) and using images that were held-out from the training dataset. Because the number of categories per tree level, and therefore chance level accuracy, differ, we normalize classification accuracy by each tree level's chance accuracy. Our final aggregate metric is $\sum_{i=1}^{n} \log\frac{a_{i}}{c_{i}}$, where $a_{i}$ is the accuracy at tree level $i$, $c_{i}$ is chance accuracy at tree level $i$, and $n$ is the number of tree levels (Figure~\ref{fig:imagenet}B). Individual accuracies across tree levels are shown in Supplementary Figure~\ref{fig:tree_level_acc}. The model trained with generative similarity had overall the highest accuracy in predicting category across WordNet tree levels. 

To show the ability of GenSim to predict human behavior in processing natural images at different levels of abstraction, we used the Levels Dataset of \citet{muttenthaler2024improving} (Figure~\ref{fig:imagenet}C). In this dataset, humans perform an odd-one-out judgment on image triplets that are sampled at three different levels of abstraction. In ``fine-grained'' triplets, all images are from the same ImageNet category. In ``coarse-grained'', they are from different categories. In ``class-boundary'', two images are from the same category and one image is from a different category. For each trial when the participant chooses an option $a$ out of three images $a,b,c$, we use our models' embeddings to compute $softmax[s(a,b),s(a,c),s(b,c)]$ where $s$ is the cosine similarity between the respective images' model embeddings. We then report $s(b,c)$ as the model's likelihood for the participant's choice (since the participant determined $a$ is the odd-one-out). The softmax temperature was choosen as $\tau=\alpha*rt$, where $rt$ is the amount of time the human participant used to make a decision and $\alpha$ is a subject-specific parameter we gridsearch over for each human participant to maximize the mean likelihood over choices across trials. We fit a seperate $\alpha$ for each participant and each \textit{model} to ensure a fair comparison across models. Our results, shown in Figure~\ref{fig:imagenet}D, indicate that the GenSim model has the highest likelihood for human choices across all three triplet types. 

\section{Discussion}\label{discussion}
We have introduced a new framework for learning human-aligned representations by combining a Bayesian notion of generative similarity with contrastive learning. To demonstrate the utility and flexibility of our approach we applied it to an array of domains relevant to human intelligence, including representing shape regularity (Figure~\ref{fig:quads}), few-shot generalization of abstract Euclidean geometry concepts (Figure~\ref{fig:programs}), and encoding of hierarchical structure in natural image categories (Figure~\ref{fig:imagenet}). 


The main limitation of our work is the need to specify a generative model to train a model with generative similarity. An alternative ``model-free'' approach is to use human data, such as similarity judgments \cite{muttenthaler2024aligning,muttenthaler2024improving}, to learn human-aligned representations. However, our method may be preferred when it is infeasible to collect a sufficiently large dataset of human judgments for training large machine learning models. This trade-off mirrors a broader trend in machine learning: human data provides direct alignment but costs resources to collect, while using structured prior knowledge requires much less resources at the cost of domain-specific assumptions. In terms of circumventing the challenges of collecting large-scale human data by leveraging structured prior knowledge, this is reminiscent of post-training of Large Language Models (LLMs), where reinforcement learning with human feedback was initially used to align models \cite{christiano2017deep}. However, the high cost of scaling human feedback datasets led to the adoption of reward functions which leverage ground-truth models for verifiable, known behaviors (e.g., math proofs or coding problems) to generate feedback on reasoning traces and produce LLMs with reasoning capabilities \cite{guo2025deepseek}. Beyond the topic of human alignment, there are other scientific domains, such as Genomics or Ecology, where large-scale human annotations are impractical, but domain expertise can inform Bayesian generative models. Such domains may also benefit from our framework.

Although we focused on domains related to vision, our framework is general enough to be applicable for other modalities. For example, LLMs can often produce unpredictable failures in logical \citep{wan2024b} and causal \citep{kiciman2023causal} reasoning. Cognitive scientists have written models for human logical reasoning \citep{piantadosi2016logical} or causal learning \citep{goodman2011learning} based on probabilistic Bayesian inference. Although contrastive learning is most commonly used in vision, there is precedence for its use in the language domain \citep{gao2021simcse,luo2024moelora}. Therefore, potential future work may involve contrastive learning with generative similarity over Bayesian models of human reasoning to imbue language models with human-like logical and causal reasoning.


\section{Acknowledgements}
This work was supported by grant N00014-23-1-2510 from the Office of Naval Research. S.K. is supported by a Google PhD Fellowship. We thank Mathias Sable-Meyer for assisting us with accessing the data in his work and general advice.
\bibliography{main}
\bibliographystyle{icml2025}
\newpage
\appendix
\onecolumn
\counterwithin*{equation}{section}
\renewcommand\theequation{\thesection\arabic{equation}}
\renewcommand{\thefigure}{S\arabic{figure}}
\renewcommand{\theHfigure}{S\arabic{figure}} 

\section*{Appendix}

\setcounter{figure}{0} 

\section{Triplet Loss Formulation of Generative Similarity} \label{app:triplet}

Given a generative model of the data,
we can define a corresponding \textit{contrastive} generative model
on data triplets $(X,X^{+},X^{-})$ as follows
\begin{align}
p_{c}(x,x^{+},x^{-})= 
\int p(x|\theta^{+})p(x^{+}|\theta^{+})p(x^{-}|\theta^{-})p(\theta^{+})p(\theta^{-})d\theta^{+}d\theta^{-}\label{eq:congendist}
\end{align}
and then given a choice of a triplet contrast function $\ell$, e.g., $d(\phi(x),\phi(x^+))-d(\phi(x),\phi(x^-))$ or some monotonic function of it (alternatively, one could also use an embedding similarity measure such as the dot product; \citealp{sohn2016improved}), we define the optimal embedding to be
\begin{equation}
\varphi^{\ast}=\arg\min_{\varphi}\mathbb{E}_{p_{c}}\ell(\phi_\varphi(X),\phi_\varphi(X^{+}),\phi_\varphi(X^{-}))\label{eq:loss}
\end{equation}
Crucially, this function can easily be estimated via Monte Carlo with triplets sampled from \\
\begin{equation}\label{eq:monte-carlo}
\begin{cases}
\theta^{+},\theta^{-}\sim p(\theta), & \text{sample distributions}\\
x,x^{+}\sim p(x|\theta^{+}), & \text{sample `same' examples}\\
x^{-}\sim p(x|\theta^{-}), & \text{sample `different' examples}
\end{cases}
\end{equation}

The functional in \Eqref{eq:loss} has a desirable property: in Appendix \ref{app:separation} we prove that
if $\ell$ is chosen to be convex and strictly increasing in $\Delta_\phi(x,x^+,x^-)\equiv d(\phi(x),\phi(x^+))-d(\phi(x),\phi(x^-))$ (e.g., softmax loss $\ell(\Delta_\phi)=\log(1+\exp(\Delta_\phi))$; \citealp{sohn2016improved}), then the optimal embedding that minimizes \Eqref{eq:loss} ensures that the expected distance between same pairs is strictly smaller than that of different pairs as defined by the processes in Figure~\ref{fig:schematic}A, i.e.,
\begin{equation}\label{eq:sep-result}
\mathbb{E}_{p_{\text{same}}}d(\phi^{\ast}(X),\phi^{\ast}(X^{+})) < \mathbb{E}_{p_{\text{diff}}}d(\phi^{\ast}(X),\phi^{\ast}(X^{-})).
\end{equation} 

Note that, for cases where the generative model is a mixture of discrete categories (usually human-labeled), where one samples a random category and then a random image within that category, this triplet loss becomes equivalent to SupCon \cite{khosla2020supervised}. This formulation has been used as a baseline for the main experiments of this work. In addition to accomodating SupCon, this framework is also flexible enough to account for more complex generative models. 

\section{Separation in Expectation}\label{app:separation}
Our goal is to show that for any triplet loss function $\ell(\Delta_\phi)$ that is convex and strictly increasing in $\Delta_\phi(x,x^+,x^-)=d_\phi(x,x^+)-d_\phi(x,x^-)$ where $d_\phi(x,y)=d(\phi(x),\phi(y))$ is a given embedding distance measure (e.g., softmax loss or quadratic loss), then the optimal embedding that minimizes \Eqref{eq:loss} ensures that the expected distance between same pairs is strictly smaller than that of different pairs as defined by the generative process in Figure \ref{fig:schematic}A. To see that, let $\phi^\ast$ denote the optimal embedding and let $\ell^\ast$ denote its achieved loss. By definition, for any suboptimal embedding $\phi_\text{sub}$ which achieves $\ell_\text{sub}$ we have $\ell^\ast<\ell_\text{sub}$. One such suboptimal embedding (assuming non-degenerate distributions) is the constant embedding which collapses all samples into a point $\phi_\text{sub}=\phi_0$. In that case, we have $\Delta=0$ and hence $\ell^\ast<\ell(0)$. Now, using Jensen inequality we have
\begin{equation}
\ell(0)>\ell^\ast=\mathbb{E}_{p_c}\ell(\Delta_{\phi^\ast}(X,X^+,X^-))\geq \ell(\mathbb{E}_{p_c}\Delta_{\phi^\ast}(X,X^+,X^-)).
\end{equation}
Observe next that since $\ell$ is strictly increasing (and hence its inverse is well-defined and strictly increasing) it follows that $\mathbb{E}_{p_c}\Delta_{\phi^\ast}(X,X^+,X^-)<0$. Finally, by noting that
\begin{align*}
\mathbb{E}_{p_{c}}d_{\phi^{\ast}}(X,X^{+}) & =\int p(x|\theta^{+})p(x^{+}|\theta^{+})p(x^{-}|\theta^{-})p(\theta^{+})p(\theta^{-})d_{\phi^{\ast}}(x,x^{+})dxdx^{+}dx^{-}d\theta^{+}d\theta^{-}\\
 & =\int p(x|\theta^{+})p(x^{+}|\theta^{+})p(\theta^{+})d_{\phi^{\ast}}(x,x^{+})dxdx^{+}d\theta^{+}\\
 & =\mathbb{E}_{p_{\text{same}}}d_{\phi^{\ast}}(X,X^{+})
\end{align*}
and likewise,
\begin{align*}
\mathbb{E}_{p_{c}}d_{\phi^{\ast}}(X,X^{-}) & =\int p(x|\theta^{+})p(x^{+}|\theta^{+})p(x^{-}|\theta^{-})p(\theta^{+})p(\theta^{-})d_{\phi^{\ast}}(x,x^{-})dxdx^{+}dx^{-}d\theta^{+}d\theta^{-}\\
 & =\int p(x|\theta^{+})p(x^{-}|\theta^{-})p(\theta^{+})p(\theta^{-})d_{\phi^{\ast}}(x,x^{-})dxdx^{-}d\theta^{+}d\theta^{-}\\
 & =\mathbb{E}_{p_{\text{diff}}}d_{\phi^{\ast}}(X,X^{-})
\end{align*}
we arrive at the desired result
\begin{equation}\label{eq:app_sep}
\mathbb{E}_{p_{\text{same}}}d_{\phi^{\ast}}(X,X^{+}) < \mathbb{E}_{p_{\text{diff}}}d_{\phi^{\ast}}(X,X^{-}).
\end{equation}

\section{Generative Similarity as an Optimal Solution}\label{app:opt-gen-sim}
In what follows we will show that generative similarity (\Eqref{eq:gen_sim}) can be derived as the minimizer of the following functional
\begin{equation}
\mathcal{L}[s]=D_{KL}[s||p_{\text{same}}]-D_{KL}[s||p_{\text{diff}}]-\beta^{-1}H(s)+\lambda\left[\int s(x,x^{\prime})dxdx^{\prime}-1\right]\label{eq:alternative-loss}
\end{equation}
where $D_{KL}$ is the Kullback-Leibler divergence, $H$ is entropy,
and the linear integral is a Lagrangian constraint that ensures that $s$ is normalized so that the other terms are well defined. Note that while
$\lambda>0$ is a Lagrange multiplier, $\beta^{-1}>0$ is a free parameter
of our choice that controls the contribution of the entropy term and
we may set it to one or a small number if desired. In other words,
the minimizer $s^{\ast}$ of $\mathcal{L}$ is the maximum-entropy
(or entropy-regularized) solution that maximizes the contrast between
$p_{\text{same}}$ and $p_{\text{diff}}$ in the $D_{KL}$ sense
(i.e. it seeks to assign high weight to pairs with high $p_{\text{same}}$
but low $p_{\text{diff}}$, and low values for pairs with low
$p_{\text{same}}$ but high $p_{\text{diff}}$). To derive $s^\ast$, observe that from the definition of the KL divergence
we have
\begin{align*}
D_{KL}[s||p_{\text{same}}]-D_{KL}[s||p_{\text{diff}}] & =\int \left[s(x,x^{\prime})\log\frac{s(x,x^{\prime})}{p_{\text{same}}(x,x^{\prime})}-s(x,x^{\prime})\log\frac{s(x,x^{\prime})}{p_{\text{diff}}(x,x^{\prime})}\right]dxdx^{\prime}\\
 & =\int \left[s(x,x^{\prime})\log p_{\text{diff}}(x,x^{\prime})-s(x,x^{\prime})\log p_{\text{same}}(x,x^{\prime})\right]dxdx^{\prime}
\end{align*}
Next, varying the functional with respect to $s$ we
have
\begin{equation}
\frac{\delta\mathcal{L}}{\delta s}=\log p_{\text{diff}}(x_{1},x_{2})-\log p_{\text{same}}(x_{1},x_{2})+\beta^{-1}\left[\log s(x_{1},x_{2})+1\right]+\lambda=0
\end{equation}
which yields
\begin{equation}
s^{\ast}(x_{1},x_{2})=\frac{1}{Z(\beta,\lambda)}\left[\frac{p_{\text{same}}(x_{1},x_{2})}{p_{\text{diff}}(x_{1},x_{2})}\right]^{\beta}
\end{equation}
\\
where we defined $Z(\beta,\lambda)\equiv e^{\beta\lambda+1}$. Next,
from the Lagrange multiplier equation $\delta_{\lambda}\mathcal{L}=0$
we have
\begin{equation}
Z(\beta,\lambda)=\int \left[\frac{p_{\text{same}}(x_{1},x_{2})}{p_{\text{diff}}(x_{1},x_{2})}\right]^{\beta}dx_{1}dx_{2}
\end{equation}
which fixes $\lambda$ as a function of $\beta$ assuming that the
right-hand integral converges. Two possible sources of divergences
are i) $p_{\text{diff}}(x_{1},x_{2})$ approaches zero while
$p_{\text{same}}(x_{1},x_{2})$ remains finite, and ii) the integral
is carried over an unbounded region without the ratio decaying fast
enough. The latter issue can be resolved by simply assuming that the
space is large but bounded and that the main probability mass of the
generative model is far from the boundaries (which is plausible for practical
applications). As for the former, observe that when $p_{\text{diff}}(x_{1},x_{2})\equiv\int p(x_{1}|\theta_{1})p(x_{2}|\theta_{2})p(\theta_{1})p(\theta_{2})d\theta_{1}d\theta_{2}=0$
it implies (from non-negativity) that $p(x_{1}|\theta_{1})p(x_{2}|\theta_{2})=0$
for all $\theta_{1,2}$ in the support of $p(\theta)$ which in turn implies that $p_{\text{same}}(x_{1},x_{2})\equiv\int  p(x_{1}|\theta)p(x_{2}|\theta)p(\theta)d\theta=0$.
In other words, if $p_{\text{different}}$ vanishes then so does $p_{\text{same}}$
(but not vice versa, e.g. if $p(x_{1}|\theta)$ and $p(x_{2}|\theta)$
have non-overlapping support as a function of $\theta$). Likewise,
the rate at which these approach zero is also controlled by the same
factor $p(x_{1}|\theta_{1})p(x_{2}|\theta_{2})\rightarrow0$ and so
we expect the ratio to be generically well-behaved. 

Finally, setting
$\beta=1$ we arrive at the desired Bayes odds relation
\begin{equation}
s^{\ast}(x_{1},x_{2})\propto\frac{p_{\text{same}}(x_{1},x_{2})}{p_{\text{diff}}(x_{1},x_{2})}=\frac{p(\text{same}|x_{1},x_{2})}{p(\text{different}|x_{1},x_{2})}
\end{equation}
where the second equality follows from the fact that we assumed that \textit{a priori} $p(\text{same})=p(\text{different})$. As a sanity check of the convergence assumptions, consider the case of a mixture of two one-dimensional Gaussians with means
$\mu_{1}=-\mu_{2}=\mu$ and uniform prior, and as a test let
us set $\sigma=1$ and $\mu\gg1$ so that the Gaussians
do not overlap and are far from the origin. In this case,
we assume that the space is finite $x\in[-\Lambda,\Lambda]$ such
that $\Lambda\gg\mu\gg1$ so that the Gaussians are unaffected by
the boundary. Then, for points that are far from the Gaussian centers,
e.g. at the origin $x_{1}=x_{2}=0$ for which the likelihoods are
exponentially small we have
\begin{equation}
\frac{p(x_{1},x_{2}|\text{same})}{p(x_{1},x_{2}|\text{different})}=\frac{\frac{1}{2}e^{-\frac{(+\mu)^{2}}{2}}\times e^{-\frac{(+\mu)^{2}}{2}}+\frac{1}{2}e^{-\frac{(-\mu)^{2}}{2}}\times e^{-\frac{(-\mu)^{2}}{2}}}{(\frac{1}{2}e^{-\frac{(+\mu)^{2}}{2}}+\frac{1}{2}e^{-\frac{(-\mu)^{2}}{2}})\times(\frac{1}{2}e^{-\frac{(+\mu)^{2}}{2}}+\frac{1}{2}e^{-\frac{(-\mu)^{2}}{2}})}=1
\end{equation}
which is indeed finite.

\section{Connections to Other Loss Functions}\label{app:otherloss}
The unregularized divergence difference $D_{KL}[s||p_{\text{same}}]-D_{KL}[s||p_{\text{diff}}]$ can also be related to a special case of the loss objective in \Eqref{eq:loss}. Specifically, observe that
\begin{equation}\label{eq:div}
D_{KL}[s||p_{\text{same}}]-D_{KL}[s||p_{\text{diff}}]=\int [s(x,x^{\prime})\log p_{\text{diff}}(x,x^{\prime})-s(x,x^{\prime})\log p_{\text{same}}(x,x^{\prime})]dxdx^{\prime}
\end{equation}
where $D_{KL}[p||q]=\int  p(x)\log[p(x)/q(x)]dx$ is the Kullback-Leibler (KL) divergence. While the left-hand-side in \Eqref{eq:div} may seem rather different from \Eqref{eq:loss}, the cancellation in the KL divergences yields a special case of \Eqref{eq:loss} with $\ell(\Delta)=\Delta$ upon minimal redefinitions, namely, recasting distance measures as similarities $d(x,y)\rightarrow s_0-s(x,y)$ and substituting probabilities with their logarithm $p\rightarrow\log p$ (i.e., applying a monotonic transformation; see Appendix \ref{app:lin}). Indeed, varying the functional in \Eqref{eq:loss} with respect to $s$ along with a simple quadratic regularizer (see Appendix \ref{app:lin}) yields $s^\ast(x_1,x_2)\propto p_\text{same}(x_1,x_2)-p_\text{diff}(x_1,x_2)$ which is equivalent to the generative similarity measure (\Eqref{eq:gen_sim}) up to a monotonic transformation of probabilities $p\rightarrow\log p$. 

\section{The Special Case of $\ell(\Delta)=\Delta$}\label{app:lin}
We consider the triplet loss objective under the special case of $\ell(\Delta)=\Delta$ where $\Delta(x,x^+,x^-)=d(x,x^+)-d(x,x^-)$. Recasting the distance measures as similarities $d(x,y)\rightarrow s_0-s(x,y)$ and unpacking \Eqref{eq:loss} we have
\begin{align*}
\mathcal{L}[s]&=
\mathbb{E}_{p_c}\Delta(X,X^+,X^-)\\&=\mathbb{E}_{p_c}s(X,X^-)-\mathbb{E}_{p_c}s(X,X^+)\\
&=\mathbb{E}_{p_\text{diff}}s(X,X^-)-\mathbb{E}_{p_\text{same}}s(X,X^+)\\
&=\int [s(x,x^{\prime})p_{\text{diff}}(x,x^{\prime})-s(x,x^{\prime}) p_{\text{same}}(x,x^{\prime})]dxdx^{\prime}
\end{align*}
where the third equality follows from an identical derivation to the one found in Appendix \ref{app:separation} above \Eqref{eq:app_sep}. 

Our goal next is to find the similarity function which minimizes $\mathcal{L}[s]$ by varying it with respect to $s$, i.e., $\delta_s\mathcal{L}=0$. As before, since $\mathcal{L}$ is linear in $s$ we need to add a suitable regularizer to derive a solution (otherwise $\delta_s\mathcal{L}=0$ has no solutions). Here we are no longer committed to a probabilistic interpretation of $s$ and so a natural choice would be a quadratic regularizer
\begin{equation}
\mathcal{L}_\text{reg}[s]=\mathcal{L}[s]+\lambda\left(\int s^{2}(x,x^{\prime})dxdx^{\prime}-\Lambda\right)
\end{equation}
for some constants $\Lambda,\lambda>0$. Varying the Lagrangian with respect to the similarity measure we have
\begin{equation}
\frac{\delta\mathcal{L_\text{reg}}}{\delta s}=p_{\text{diff}}(x_{1},x_{2})-p_{\text{same}}(x_{1},x_{2})+2\lambda s(x_1,x_2)=0
\end{equation}
This in turn implies that the optimal similarity measure is given
by
\begin{equation}
s^{\ast}(x_{1},x_{2})=\frac{1}{2\lambda}\left[p_{\text{same}}(x_{1},x_{2})-p_{\text{diff}}(x_{1},x_{2})\right]
\end{equation}
Likewise, for the Lagrange multiplier we have
\begin{equation}
\frac{\delta\mathcal{L}_\text{reg}}{\delta\lambda}=\int s^{2}(x,x^{\prime})dxdx^{\prime}-\Lambda=0
\end{equation}
Plugging in the optimal solution we have
\begin{equation}
\frac{1}{4\lambda^{2}}\int \left[p_\text{same}(x,x^{\prime})-p_\text{diff}(x,x^{\prime}))\right]^{2}dxdx^{\prime}-\Lambda=0
\end{equation}
The integral is positive since it is the squared difference
between two normalized probability distributions, and so denoting its value as $C_{p}>0$ we
can solve for $\lambda$
\begin{equation}
\lambda=\frac{1}{2}\sqrt{\frac{C_{p}}{\Lambda}}
\end{equation}
Thus, putting everything together we have
\begin{equation}
s^{\ast}(x_{1},x_{2})=\sqrt{\frac{\Lambda}{C_{p}}}\left[p_\text{same}(x_1,x_2)-p_\text{diff}(x_1,x_2))\right]\label{eq:final-optimal-sim}
\end{equation}

\section{Illustrative Example: Generative Similarity of a Gaussian Mixture}\label{app:gaussian_triplet_exp}

\begin{figure}[h]
  \centering  \includegraphics[width=\linewidth]{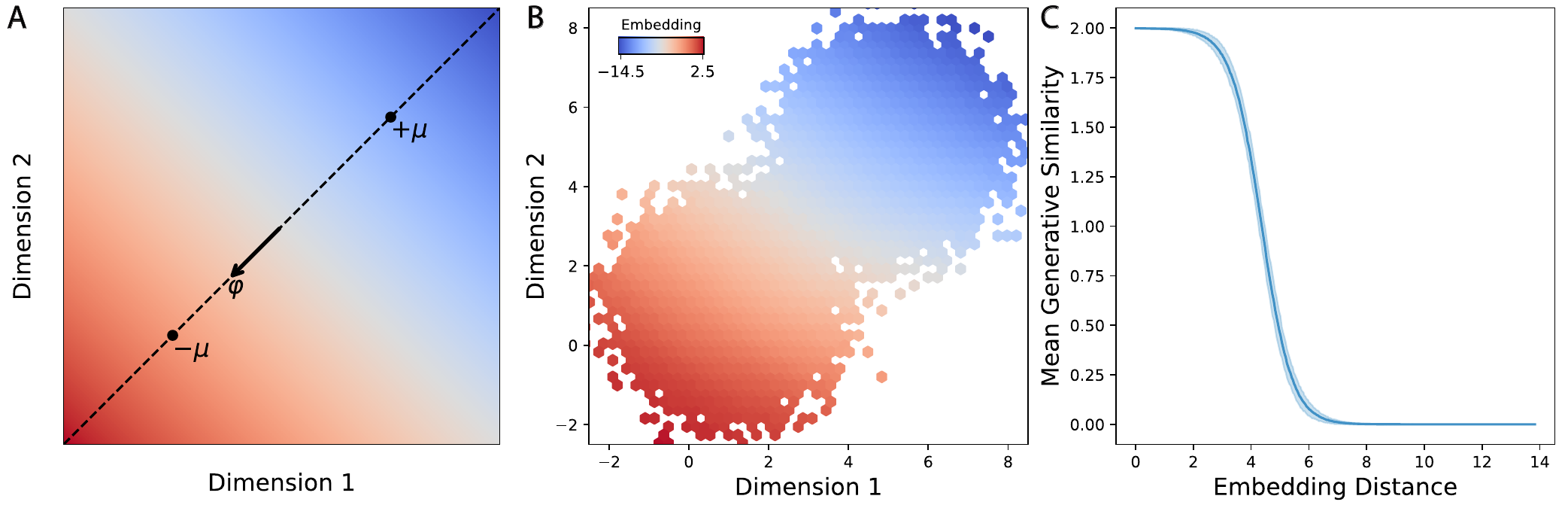}
  \caption{\textbf{Encoding the generative similarity of a Gaussian mixture.} \textbf{A.} Optimal linear projection vector $\varphi$ for a symmetric Gaussian mixture with means $\pm\mu$. \textbf{B.} Learned 1D embedding values from a two-layer perceptron for points sampled from a 2D Gaussian mixture (colors indicate values). \textbf{C.} Mean generative similarity as a function of distance in the embedding space shown in \textbf{B} (discretized into 500 quantile bins). Shaded area indicates 95\% CIs bootsrapped over data points.}
  \label{fig:gaussian-simulation}
\end{figure}

To get a sense of the contrastive training procedure with generative similarity, we consider a Gaussian mixture example. Gaussian mixtures are an ideal starting point because i) they are analytically and numerically tractable, and ii) they play a key role in the cognitive literature on models of categorization \citep{rosseel2002mixture,sanborn2010rational}.
Consider a data generative process that is given by a mixture of two
Gaussians with means $\mu_{1,2}$, equal variances $\sigma^2$,
and a uniform prior $p_{1,2}=1/2$. Without loss of generality,
we can choose a coordinate system in which $\mu_{1}=-\mu_{2}\equiv\mu$.
Consider further a subfamily of embeddings that are specified by linear projections $\phi_\varphi(x)=\varphi\cdot x$
where $\varphi$ is a unit vector of choice $\varphi\cdot\varphi=1$. A natural measure of contrastive loss in this case would be
\begin{equation}\label{eq:linloss}
\mathbb{E}_{p_c}\ell=\mathbb{E}_{p_c}\left[\phi_\varphi(X)\cdot\phi_\varphi(X^-)-\phi_\varphi(X)\cdot\phi_\varphi(X^+)\right]
\end{equation}
i.e., using the `dot' product as a measure of embedding similarity \citep{sohn2016improved} (for one-dimensional embeddings this is just a regular product). By plugging in the definition of the Gaussian mixture and simplifying using Gaussian moments (see Appendix \ref{app:gauss}), \Eqref{eq:linloss} boils down to $\mathbb{E}_{p_c}\ell\propto-4||\mu||_2^2\cos^2\theta_{\varphi\mu}$
where $\theta_{\varphi\mu}$ is the angle between $\varphi$ and $\mu$. This loss is minimized for $\cos\theta_{\varphi\mu}=\pm1$ or equivalently $\varphi^\ast=\pm\hat{\mu}$ where
$\hat{\mu}$ is the normalized version of $\mu$. This means that
the optimal linear mapping is simply one that projects different points onto
the axis connecting the centers of the two Gaussians $\mu_{1}-\mu_{2}=2\mu$,
which is equivalent to a linear decision boundary that is orthogonal
to the line $\mu_{1}-\mu_{2}$ and passing through the origin, and thus effectively recovering a linear classifier (Figure~\ref{fig:gaussian-simulation}A). We further confirmed in a simulation that this behavior persists when applying the Monte-Carlo approximation in \Eqref{eq:monte-carlo} with a quadratic loss to a two-layer perceptron (Figure~\ref{fig:gaussian-simulation}B; classification test accuracy of $99.7\%$; additional details in Appendix \ref{app:gauss-sim}). Moreover, we compared distance in the learned embedding space to the theoretical generative similarity values in that case and found excellent agreement (Spearman's $\rho(498)=-0.99$, $p<10^{-3}$; Figure~\ref{fig:gaussian-simulation}C). We also found that the 95\% confidence interval (CI) on the average distance for same pairs was $[1.62,1.66]$ whereas for different pairs it was $[4.77,4.91]$, consistent with the prediction of \Eqref{eq:sep-result}. 

\section{Gaussian Mixtures and Linear Projections}\label{app:gauss}
To derive the linear projection result, we start by plugging in the definition of the Gaussian mixture generative process (i.e., uniformly sampling a Gaussian and then sampling points from it) into \Eqref{eq:loss}
\begin{align*}
\mathbb{E}_{p_{c}}\ell(X,X^{+},X^{-}) & =\sum_{i=1,2}\sum_{j=1,2}\int \ell(\phi(x),\phi(x^{+}),\phi(x^{-}))\times\\
 & \times\frac{1}{4}\frac{1}{((2\pi)^{d}\sigma^{2d})^{3/2}}\exp\left(-\frac{(x-\mu_{i})^{2}+(x^{+}-\mu_{i})^{2}+(x^{-}-\mu_{j})^{2}}{2\sigma^{2}}\right)dxdx^{+}dx^{-}
\end{align*}
Now, recall that
\begin{align*}
\ell(x,x^+,x^-)&=\phi_\varphi(x)\cdot\phi_\varphi(x^-)-\phi_\varphi(x)\cdot\phi_\varphi(x^+)\\
&=(\varphi\cdot x)(\varphi\cdot x^-)-(\varphi\cdot x)(\varphi\cdot x^+)
\end{align*}
Substituting into the loss formula and integrating we have
\begin{align*}
\mathbb{E}_{p_{c}}\ell & \propto\sum_{i,j}\int (\varphi\cdot x)(\varphi\cdot x^-)\exp\left(-\frac{(x-\mu_{i})^{2}+(x^{-}-\mu_{j})^{2}}{2\sigma^{2}}\right)dxdx^{-}\\
 & -2\sum_{i}\int (\varphi\cdot x)(\varphi\cdot x^+)\exp\left(-\frac{(x-\mu_{i})^{2}+(x^{+}-\mu_{i})^{2}}{2\sigma^{2}}\right)dxdx^{+}
\end{align*}
Note next that since $x^{+}$ and $x^{-}$ are dummy integration variables,
we can further rewrite
\begin{align*}
\mathbb{E}_{p_{c}}\ell & \propto\sum_{i\ne j}\int (\varphi\cdot x)(\varphi\cdot x^-)\exp\left(-\frac{(x-\mu_{i})^{2}+(x^{-}-\mu_{j})^{2}}{2\sigma^{2}}\right)dxdx^{-}\\
 & -\sum_{i}\int (\varphi\cdot x)(\varphi\cdot x^+)\exp\left(-\frac{(x-\mu_{i})^{2}+(x^{+}-\mu_{i})^{2}}{2\sigma^{2}}\right)dxdx^{+}
\end{align*}
Thus, using the fact that the distributions are separable and standard Gaussian moment formulae we arrive at
\begin{equation}
\mathbb{E}_{p_c}\ell\propto\sum_{i\ne j}(\varphi\cdot\mu_i)(\varphi\cdot\mu_j)-\sum_i(\varphi\cdot\mu_i)(\varphi\cdot\mu_i)
\end{equation}
Finally, plugging in $\mu_1=-\mu_2=\mu$ and using the fact that $||\varphi||^2_2=\varphi\cdot\varphi=1$ we have
\begin{equation}
\mathbb{E}_{p_c}\propto-4(\varphi\cdot\mu)^2=-4||\mu||_2^2\cos^2\theta_{\varphi\mu}.
\end{equation}

\section{Gaussian Mixtures and Two-Layer Perceptrons}\label{app:gauss-sim}
To test the Monte Carlo approximation of \Eqref{eq:monte-carlo} and to see how well it tracks the theoretical generative similarity, we considered an embedding family that is parametrized by two-layer perceptrons. For the generative family, we chose as before a mixture of two Gaussians, this time with mean values of $\mu_1=(5,5)$ and $\mu_2=(1,1)$ and unit variance $\sigma^2=1$. As for the loss function, here we used a quadratic (Euclidean) loss of the form
\begin{equation}
    \mathcal{L}=\frac{1}{N_\text{triplets}}\sum_{\{x,x^+,x^-\}}\left[(\phi(x)-\phi(x^+))^2-(\phi(x)-\phi(x^-))^2\right]
\end{equation}
where $\{x,x^+,x^-\}$ are triplets sampled from \Eqref{eq:monte-carlo}. We trained the perceptron model using 10,000 triplets (learning rate $= 10^{-5}$, hidden layer size $= 32$, batch-size $=256$, and $300$ epochs). The resulting model successfully learned to distinguish the two Gaussians as seen visually from the embedding values in Figure~\ref{fig:gaussian-simulation}B, and also from the test accuracy of $99.7\%$. Finally, we wanted to see how well the embedding distance tracked the theoretical generative similarity, which in this case can be derived in closed form by simply plugging in the Gaussian distributions in \Eqref{eq:gen_sim}
\begin{equation}
s(x_{1},x_{2})=\frac{\frac{1}{2}e^{-\frac{(x_{1}-\mu_{1})^{2}}{2\sigma^{2}}}\times e^{-\frac{(x_{2}-\mu_{1})^{2}}{2\sigma^{2}}}+\frac{1}{2}e^{-\frac{(x_{1}-\mu_{2})^{2}}{2\sigma^{2}}}\times e^{-\frac{(x_{2}-\mu_{2})^{2}}{2\sigma^{2}}}}{(\frac{1}{2}e^{-\frac{(x_{1}-\mu_{1})^{2}}{2\sigma^{2}}}+\frac{1}{2}e^{-\frac{(x_{1}-\mu_{2})^{2}}{2\sigma^{2}}})\times(\frac{1}{2}e^{-\frac{(x_{2}-\mu_{1})^{2}}{2\sigma^{2}}}+\frac{1}{2}e^{-\frac{(x_{2}-\mu_{2})^{2}}{2\sigma^{2}}})}.
\end{equation}
The mean generative similarity as a function of embedding distance between pairs (grouped into 500 quantile bins) is shown in Figure~\ref{fig:gaussian-simulation}C. We see that it is indeed a monotonically decreasing function of distance in embedding space (Spearman's $\rho(498)=-0.99$, $p<10^{-3}$). Moreover, we found that the average distance 95\% (1.96-sigma) confidence interval (CI) for same pairs was $[1.62,1.66]$ whereas for different pairs it was $[4.77,4.91]$, consistent with the prediction of \Eqref{eq:sep-result}.

\section{Generative Similarity of Geometric Shape Distributions}\label{app:shape-deriv}
Our goal is to derive the generative similarity measure associated with the process in \Eqref{eq:shape-process}
\begin{equation}
s(\sigma_{1}(f^{(1)}),\sigma_{2}(f^{(2)}))=\frac{p_\text{same}(\sigma_{1},\sigma_{2})}{p_\text{diff}(\sigma_{1},\sigma_{2})}
\end{equation}
Plugging in the different Beta, Bernoulli, and uniform distributions into the nominator of \Eqref{eq:gen_sim} we have
\begin{align*}
p_\text{same}(\sigma_{1},\sigma_{2})&=\sum_{\hat{f}_{1\cdots n}^{(1,2)}}\int \frac{\delta(\hat{f}^{(1)}-f^{(1)})}{|\mathcal{S}(\hat{f}^{(1)})|}\frac{\delta(\hat{f}^{(2)}-f^{(2)})}{|\mathcal{S}(\hat{f}^{(2)})|}\\
&\times\prod_{i}\left[\theta_{i}^{f_{i}^{(1)}+f_{i}^{(2)}+\alpha-1}(1-\theta_{i})^{\bar{f}_{i}^{(1)}+\bar{f}_{i}^{(2)}+\beta-1}/\text{B}(\alpha,\beta)\right]d\theta_{1\cdots n}
\end{align*}
where we defined $\bar{f}_{i}=1-f_{i}$ and used the definition of
the Bernoulli distribution $\text{Bern}(f_i;\theta_i)=\theta_i^{f_i}(1-\theta_i)^{\bar{f}_i}$, and the Beta distribution $\text{Beta}(\theta_{i};\alpha,\beta)=\theta_{i}^{\alpha-1}(1-\theta_{i})^{\beta-1}/\text{B}(\alpha,\beta)$
where $\text{B}$ is the Beta function and is given by $\text{B}(z_{1},z_{2})=\int_{0}^{1}dtt^{z_{1}-1}(1-t)^{z_{2}-1}$
which is well-defined for all positive numbers $z_{1},z_{2}>0$. Note that the delta function $\delta(\hat{f}-f)$ simply enforces the fact that by definition each stimulus is consistent with only one set of feature values (otherwise there would be at least one feature of the stimulus that is both True and False which is a contradiction). Likewise, $|\mathcal{S}(\hat{f})|$ is the cardinality of the exemplar set associated with the feature vector $\hat{f}$ which accounts for uniform sampling. Likewise, for the denominator of \Eqref{eq:gen_sim} we have
\begin{align*}
p_\text{diff}(\sigma_{1},\sigma_{2}) & =\sum_{\hat{f}_{1\cdots n}^{(1)}}\int \frac{\delta(\hat{f}^{(1)}-f^{(1)})}{|\mathcal{S}(\hat{f}^{(1)})|}\prod_{i}\left[\theta_{(1)i}^{f_{i}^{(1)}+\alpha-1}(1-\theta_{(1)i})^{\bar{f}_{i}^{(1)}+\beta-1}/\text{B}(\alpha,\beta)\right]d\theta_{1\cdots n}^{(1)}\\
 & \times\sum_{\hat{f}_{1\cdots n}^{(2)}}\int \frac{\delta(\hat{f}^{(2)}-f^{(2)})}{|\mathcal{S}(\hat{f}^{(2)})|}\prod_{j}\left[\theta_{(2)j}^{f_{j}^{(2)}+\alpha-1}(1-\theta_{(2)j})^{\bar{f}_{j}^{(2)}+\beta-1}/\text{B}(\alpha,\beta)\right]d\theta_{1\cdots n}^{(2)}\nonumber 
\end{align*}

The above integrals might seem quite complicated at first but the conjugacy relation between the Beta and Bernoulli distributions as well as the delta functions simplify things drastically. Indeed, the delta functions cancel the summation over features, and the cardinality factors cancel out in the ratio so that we are left with a collection of Beta function factors (see definition of Beta function above)
\begin{equation}
s(\sigma_{1}(f^{(1)}),\sigma_{2}(f^{(2)}))=\frac{\prod_{i}\text{B}(f_{i}^{(1)}+f_{i}^{(2)}+\alpha,\bar{f}_{i}^{(1)}+\bar{f}_{i}^{(2)}+\beta)\text{B}(\alpha,\beta)}{\prod_{i}\text{B}(f_{i}^{(1)}+\alpha,\bar{f}_{i}^{(1)}+\beta)\prod_{j}\text{B}(f_{j}^{(2)}+\alpha,\bar{f}_{j}^{(2)}+\beta)}
\end{equation}
Taking the logarithm and rearranging the terms we have
\begin{equation}
\log s(\sigma_{1}(f^{(1)}),\sigma_{2}(f^{(2)}))=\sum_{i}\log\frac{\text{B}(f_{i}^{(1)}+f_{i}^{(2)}+\alpha,\bar{f}_{i}^{(1)}+\bar{f}_{i}^{(2)}+\beta)\text{B}(\alpha,\beta)}{\text{B}(f_{i}^{(1)}+\alpha,\bar{f}_{i}^{(1)}+\beta)\text{B}(f_{i}^{(2)}+\alpha,\bar{f}_{i}^{(2)}+\beta)}
\end{equation}
Now, recall the following Beta function identities\footnote{These follow from the fact that $\text{B}(z_1,z_2)=\Gamma(z_1)\Gamma(z_2)/\Gamma(z_1+z_2)$ where $\Gamma$ is the Gamma function which satisfies $\Gamma(z+1)=z\Gamma(z)$ for any $z>0$ \citep{artin2015gamma}.}
\begin{equation}
\text{B}(x+1,y)=\frac{x}{x+y}\text{B}(x,y);\quad\text{B}(x,y+1)=\frac{y}{x+y}\text{B}(x,y)
\end{equation}
Using these identities we can group and simplify the different ratios
contributing to the sum depending on the values of the features. If
$f_{i}^{(1)}=f_{i}^{(2)}=1$ then we have
\begin{equation}
\log\frac{\text{B}(2+\alpha,0+\beta)\text{B}(\alpha,\beta)}{\text{B}(1+\alpha,0+\beta)\text{B}(1+\alpha,0+\beta)}=\log\frac{\alpha+1}{\alpha+\beta+1}\frac{\alpha+\beta}{\alpha}
\end{equation}
If on the other hand $f_{i}^{(1)}=1$ and $f_{i}^{(2)}=0$ or $f_{i}^{(1)}=0$
and $f_{i}^{(2)}=1$ then we have
\begin{equation}
\log\frac{\text{B}(1+\alpha,1+\beta)\text{B}(\alpha,\beta)}{\text{B}(1+\alpha,0+\beta)\text{B}(0+\alpha,1+\beta)}=\log\frac{\beta}{\alpha+\beta+1}\frac{\alpha+\beta}{\beta}
\end{equation}
and finally for $f_{i}^{(1)}=f_{i}^{(2)}=0$ we have
\begin{equation}
\log\frac{\text{B}(0+\alpha,2+\beta)\text{B}(\alpha,\beta)}{\text{B}(0+\alpha,1+\beta)\text{B}(0+\alpha,1+\beta)}=\log\frac{\beta+1}{\alpha+\beta+1}\frac{\alpha+\beta}{\beta}
\end{equation}
Next, defining $\Sigma_{1}$ and $\Sigma_{2}$ to be the sets of features
that hold true for stimuli $\sigma_{1}$ and $\sigma_{2}$, we can
write
\begin{align*}
\log s(\sigma_{1}(f^{(1)}),\sigma_{2}(f^{(2)})) & =|\Sigma_{1}\cap\Sigma_{2}|\log\frac{\alpha+1}{\alpha+\beta+1}\frac{\alpha+\beta}{\alpha}\nonumber \\
 & +(|\Sigma_{1}-\Sigma_{2}|+|\Sigma_{2}-\Sigma_{1}|)\log\frac{\beta}{\alpha+\beta+1}\frac{\alpha+\beta}{\beta}\\
 & +|\bar{\Sigma}_{1}\cap\bar{\Sigma}_{2}|\log\frac{\beta+1}{\alpha+\beta+1}\frac{\alpha+\beta}{\beta}\nonumber 
\end{align*}
where $|\Sigma_{1}\cap\Sigma_{2}|$ is the number features that hold
true for both stimuli, $|\Sigma_{i}-\Sigma_{j}|$ is the number of
features that hold true for $\sigma_{i}$ but not for $\sigma_{j}$,
and finally $|\bar{\Sigma}_{1}\cap\bar{\Sigma}_{2}|$ is the number
of features that hold neither for $\sigma_{1}$ nor for $\sigma_{2}$. Observe next that by definition $|\bar{\Sigma}_{1}\cap\bar{\Sigma}_{2}|=n-|\Sigma_{1}\cap\Sigma_{2}|-|\Sigma_{1}-\Sigma_{2}|-|\Sigma_{2}-\Sigma_{1}|$
where $n$ is the overall number of features. From here it follows that
\begin{align*}
\log s(\sigma_{1}(f^{(1)}),\sigma_{2}(f^{(2)})) & =|\Sigma_{1}\cap\Sigma_{2}|\log\frac{\alpha+1}{\alpha}\frac{\beta}{\beta+1}\\
 & +(|\Sigma_{1}-\Sigma_{2}|+|\Sigma_{2}-\Sigma_{1}|)\log\frac{\beta}{\beta+1}\nonumber \\
 & +n\log\frac{\beta+1}{\alpha+\beta+1}\frac{\alpha+\beta}{\beta}\nonumber 
\end{align*}
In the limit of $\alpha=\beta\rightarrow0$ we have
\begin{equation}
\log s(\sigma_{1}(f^{(1)}),\sigma_{2}(f^{(2)}) = n\log2-\log\frac{\beta+1}{\beta}(|\Sigma_{1}-\Sigma_{2}|+|\Sigma_{2}-\Sigma_{1}|)
\end{equation}
where the first term is simply a constant. Finally, observe that
\begin{align}
|\Sigma_{1}-\Sigma_{2}|+|\Sigma_{2}-\Sigma_{1}| & =\sum_{i}f_{i}^{(1)}(1-f_{i}^{(2)})+f_{i}^{(2)}(1-f_{i}^{(1)})\\
 & =\sum_{i}f_{i}^{(1)}-2f_{i}^{(1)}f_{i}^{(2)}+f_{i}^{(2)}\nonumber \\
 & =\sum_{i}(f_{i}^{(1)}-f_{i}^{(2)})^{2}\nonumber 
\end{align}
where the third equality follows from the fact that $f^{2}=f$ for binary features. In other words, the generative similarity reduces to a monotonically decreasing function of the Euclidean distance between the geometric features of shapes
\begin{equation}
\log s(\sigma_{1}(f^{(1)}),\sigma_{2}(f^{(2)}) = n\log2-\log\frac{\beta+1}{\beta}\sum_{i}(f_{i}^{(1)}-f_{i}^{(2)})^{2}
\end{equation}
which is the desired result.

\section{Collecting Human Similarity Judgments}\label{human-exp}
We collected human similarity judgments through Prolific\footnote{\url{www.prolific.com}}, an online crowdsourcing platform. Overall, we recruited $N=214$ US participants and they were paid a fair wage of 12 USD per hour. All participants provided informed consent prior to participation in accordance with an approved institutional review board (IRB) protocol. Upon providing consent, participants received the following instructions: ``In this study, we are studying how people perceive geometric shapes. Your task is to rate the similarity between different pairs of geometric shapes. You will have seven response options, ranging from 0 ('Completely Dissimilar') to 6 ('Completely Similar'). Choose the one you think is most appropriate. Note: no prior expertise is required to complete this task, just choose what you intuitively think is the right answer.'' Participants then completed up to 100 trials. There were $66$ stimuli which corresponded to $66\times65/2=2145$ unique pairs. The order of presentation within a pair in a given trial was randomized to avoid order effects. Overall, we elicited 20,948 human judgments.

\section{Generative Similarity of Items Sampled from a Hierarchy}
\label{app:hierarchy_formula_deriv}
Our goal is to compute the generative similarity between a pair of items sampled from a hierarchy (similar to the one described in Section~\ref{im-hierarchy}). Specifically, we assume that items are sampled from a tree hierarchy by probabilistically choosing a sequence of branches from the root until a leaf is reached, after which an item is selected uniformly from a set associated with that leaf. We assume that each possible item is associated with a single leaf such that there is a unique path from the root to the relevant leaf. The probability of traversing a branch $s$ at a given depth level $k$ is specified by some value $p_{k}$. Given two items $x_i$ and $x_j$, let $K$ denote the branching or separation depth at which their trajectories from the root to their respective leaves diverge. Moreover, let $N_i$ and $N_j$ denote the depth of each respective leaf, and let $M_i$ and $M_j$ denote the size of the set of items associated with each leaf. From here, under the independent sampling assumption we can denote the probabilities of $x_i$ and $x_j$ as $p_1 p_2\dots p_K p_{K+1}^{(i)}\dots p_{N_i}^{(i)}\times\frac{1}{M_i}$ and $p_1 p_2\dots p_K p_{K+1}^{(j)}\dots p_{N_j}^{(j)}\times\frac{1}{M_j}$ for $x_i$ and $x_j$, respectively. Note that up until the branching point the probabilities are shared by definition. On the other hand, when $x_i$ and $x_j$ are sampled jointly, any joint sampling process that traverses from the root along the shared path and stops no later than the branching point is a valid joint sampling process for $x_i$ and $x_j$ and so we must sum over it. Thus, overall we have
\begin{align*}
    s(x_i,x_j)&=\frac{1}{K}\frac{1}{M_i M_j}\frac{\left[p_1(p_2\dots p_K\times p_{K+1}^{(i)}\dots p_{N_i}^{(i)})(p_2\dots p_K\times p_{K+1}^{(j)}\dots p_{N_j}^{(j)})+\dots\right]}{p_1 p_2\dots p_K p_{K+1}^{(i)}\dots p_{N_i}^{(i)}\times\frac{1}{M_i}\times p_1 p_2\dots p_K p_{K+1}^{(j)}\dots p_{N_j}^{(j)}\times\frac{1}{M_j}}\\
    &=\frac{1}{K}\left[\frac{1}{p_1}+\frac{1}{p_1 p_2}+\dots+\frac{1}{p_1p_2\cdots p_K}\right]
\end{align*}
where in the second equality we simply canceled out shared factors. This recovers \Eqref{imagenetformula} which is the desired result.

\section{Details on Quadrilateral Experiment}\label{app:quad-detail}
For our main experiments, we use the CorNet model which was used in the original work (variant `S') that introduces the Oddball task \citep{sable2021sensitivity}. CorNet contains four ``areas'' corresponding to the areas of the visual stream: V1, V2, V4, and IT. Each area contains convolutional and max pooling layers. There are also biologically plausible recurrent connections between areas (e.g., V4 to V1). After IT, the penultimate area in the visual stream, a linear layer is used to readout object categories. The model is pretrained on ImageNet-1k on a standard supervised image classification objective.

For finetuning the model on the supervised classification objective, we followed the protocol used in the supplementary results of \citet{sable2021sensitivity}. Specifically, $11$ new object categories are added to the model's last layer, and the model is trained to classify a quadrilateral as one of the $11$ categories shown in Figure~\ref{fig:quads}B. For training data, we used quadrilaterals from all $11$ categories with different scales and rotations (though the specific quadrilateral images used in the test trials were held out). We used a learning rate of 5e-6 using the Adam optimizer with a cross entropy loss. We conducted $10$ training runs. Training was conducted on an NVIDIA Quadro P6000 GPU with 25GB of memory. 

For finetuning the model on the generative similarity contrastive objective, we first calculated the Euclidean distance of the model's final layer embedding between different quadrilateral images, then calculated the Euclidean distance between the quadrilaterals' respective geometry feature vectors, and finally used the mean squared error between the embeddings' distance and the feature vectors' distance as the loss. The geometric feature vectors were a set of 22 binary features encoding the following properties: $6$ features per pair of edges encoding whether their lengths are equal or not, $6$ features per pair of angles coding whether their angles are equal or not, $6$ features per pair of edges encoding whether they were parallel or not, and $4$ features per angle encoding whether they were right angles or not. See Table~1 for a list of these values. Like the supervised model, we used training data from each category of quadrilaterals with different scales and rotations (though specific images used in the test trials were held out). We used the Adam optimizer with a learning rate of 5e-4.We conducted $10$ training runs. We used the exact same training data, learning rate, and optimizer when running the control experiments for finetuning CorNet on the SimCLR objective. For the SimCLR objective, the augmentations we used were random resize crop, random horizontal flips, and random Gaussian blurs. The original SimCLR paper also had augmentations related to color distortions that we did not use because our data were already grayscale images. Like SimCLR, we used the InfoNCE loss function \citep{oord2018representation}. Let $v_{i}$ be the embedding of image $i$ and $v_{i}^\prime$ be the embedding of image $i$'s augmented counterpart. The loss is $\frac{1}{N}\sum_{i=1}^{N}\log \frac{f(v_{i},v_{i}^\prime)}{\frac{1}{N}\sum_{j}\exp f(v_{i},v_{j}^\prime)}$ where $f$ is a similarity function between embeddings (SimCLR used cosine similarity). This effectively pushes representations of images and their augmented counterparts to be more similar while also pushing representations of images and other images' augmented counterparts to be more dissimilar. Training was conducted on an NVIDIA Quadro P6000 GPU with 25GB of memory. 
 
\begin{table}[t]
\begin{center}
\label{tablegeom}
\caption{List of geometric regularities for each quadrilateral type (sorted from most regular to least)} 
\begin{tabular}{|l|c|c|c|c|c|}
\hline
shape         & rightAngles & parallels & symmetry & equalSides & equalAngles \\ \hline
square        & 4           & 2         & 4        & 4          & 4           \\ \hline
rectangle     & 4            & 2         & 2        & 2          & 4           \\ \hline
losange       & 0           & 0         & 2        & 4          & 2           \\ \hline
parallelogram & 0           & 2         & 1        & 2          & 2           \\ \hline
rightKite     & 2           & 0         & 1        & 2          & 2           \\ \hline
kite          & 0           & 0         & 1        & 2          & 2           \\ \hline
isoTrapezoid  & 0           & 1         & 1        & 1          & 2           \\ \hline
hinge         & 1           & 0         & 0        & 1          & 0           \\ \hline
rustedHinge   & 0           & 0         & 0        & 1          & 0           \\ \hline
trapezoid     & 0           & 1         & 0        & 0          & 0           \\ \hline
random        & 0           & 0         & 0        & 0          & 0           \\ \hline
\end{tabular}
\end{center}
\end{table}

\section{Details on Geoclidean Experiment}\label{app:draw-detail}
Out of the $37$ programs or "concepts" presented in the original Geoclidean work, we randomly chose $30$ to generate our training data and the other $7$ for testing (see section \ref{app:concepts}). We generated a dataset of $100,000$ images from the training concepts. For all of our models, we finetuned the pretrained ResNet-50 architecture \cite{he2016deep} that was pretrained on ImageNet. We conducted $4$ different training runs for each condition. Each model was trained for $35$ epochs using the Adam optimizer with a batch size of 128. Images were resized to $224 \times 224 \times 3$ (for $3$ RGB channels) before training.  

For the SimCLR baseline objective, images in the batch were randomly augmented. The augmentations were: random resize crop, random horizontal flips, and random Gaussian blurs. The original SimCLR paper also had augmentations corresponding to color distortions which we did not use because our data were already grayscale images. Like SimCLR, we used the InfoNCE loss function \citep{oord2018representation}. Let $v_{i}$ be the embedding of image $i$ and $v_{i}^\prime$ be the embedding of image $i$'s augmented counterpart. The loss is $\frac{1}{N}\sum_{i=1}^{N}\log \frac{f(v_{i},v_{i}^\prime)}{\frac{1}{N}\sum_{j}\exp f(v_{i},v_{j}^\prime)}$ where $f$ is a similarity function between embeddings (SimCLR used cosine similarity). This effectively pushes representations of images and their augmented counterparts to be more similar while also pushing representations of images and other images' augmented counterparts to be more dissimilar. Training was conducted with one NVIDIA Tesla P100 GPU with 16GB of memory. Training was conducted with one NVIDIA Tesla P100 GPU with 16GB of memory. 

For the class-based baseline objective, let image $a$ be the representation of the anchor image that is sampled from a random program $c$. Let $p$ be the representation of a positive image that is another image sampled from $c$ and $n$ be the negative image that is randomly sampled from another program. Using these samples, we compute the positive Euclidean distance $d(p,a)$ and the negative Euclidean distance $d(n,a)$ in order to compute the loss function $d(p,a)-d(n,a)$. This loss is equivalent to an implementation of supervised contrastive learning \cite{khosla2020supervised}.   

For the generative similarity objective, let image $a$ be the the representation of an image that is sampled from a random program $c$ and image $b$ be another image sampled from a program $d$ (which may or may not be the same program). We compute the mean squared error between the cosine similarity between $a$ and $b$ and the Log of the tree kernel between the respective parse trees of $c$ and $d$. The tree kernel from \citet{moschitti2006making} recursively counts number of shared subtrees. Although \citet{moschitti2006making} also propose a method for subset trees, which is more exhaustive than parse trees, we found our results are sufficiently obtained through counting shared subtrees. 

\section{Details on Natural Image Hierarchy Experiment}\label{app:imagenet-detail}
Our training data came from the standard ImageNet-1k dataset (ILSVRC 2012) \citep{deng2009imagenet}. This dataset contains over a million training images from $1000$ different categories. These categories are each considered leaf nodes inside the WordNet tree hierarchy and are non-overlapping. Images were resized to $224 \times 224 \times 3$ (for $3$ RGB channels) before training. We trained a standard ResNet-50 \citep{he2016deep} architecture on all types of losses. We conducted $5$ different training runs for each condition. For all objectives, we trained for 65 epochs, we used the same learning rate 1e-3 and the same Adam optimizer with a batch size of 256. Training was conducted with one NVIDIA A100 GPU with 40GB of memory.  

For the SimCLR baseline objective, we used the same objective and image augmentations as the previous experiment, but also included random grayscaling since ImageNet images are colored. For the class-based baseline objective, we used a similar triplet-based loss as in the Geoclidean experiment, with an anchor and positive image from the same category and a negative example from a different category. For the generative similarity objective, we computed the mean-squared error between the cosine similarities between two image embeddings and their generative similarity measure as defined in Equation \ref{imagenetformula}. 

For evaluating the ability of the learned embeddings to encode information about hierarchical categories, we fit a Linear SVM on test set images to predict image categories for each tree level (i.e. we fit a different SVM for each level of the tree and each model, results for each tree level can be seen in Supplementary Figure~\ref{fig:tree_level_acc}). The SVM was fit for upto $100$ epochs (stopping when the loss fails to decrease by more than 1e-4 for $5$ epochs). We trained the SVM within a three-fold cross validation loop, evaluating classification accuracy on a held-out test set, and then reported the mean across the three folds.
\section{Geoclidean Concepts}\label{app:concepts}
Below are the $37$ programs presented in the original Geoclidean benchmark \cite{hsu2022geoclidean}. We used $30$ for training and randomly held out $7$ for testing ('cccc', 'clcc', 'quadrilateral', 'triangle', 'perp bisector', 'llcc', 'ccc'). Test concepts have a four asterisks next to their names. 
\clearpage 
\begin{verbatim}
    perp_bisector****
    'l1 = line(p1(), p2())',
    'c1* = circle(p1(), p2())',
    'c2* = circle(p2(), p1())',
    'l2 = line(p3(c1, c2), p4(c1, c2))',
    
    rectilinear
    'l1 = line(p1(), p2())',
    'l2 = line(p2(), p3())',
    'l3 = line(p3(), p4())',
    'l4 = line(p4(), p5())',
    'l5 = line(p5(), p1())',
    
    ang_bisector
    'l1 = line(p1(), p2())',
    'c1* = circle(p1(), p2())',
    'c2* = circle(p2(), p1())',
    'c3* = circle(p3(c1, c2), p1())',
    'l2 = line(p1(), p3(c1, c2))',
    'l3 = line(p1(), p4(c2, c3))',
    
    oblong
    'l1* = line(p1(), p2())',
    'c1* = circle(p1(), p2())',
    'c2* = circle(p2(), p1())',
    'l2* = line(p3(c1, c2), p4(c1, c2))',
    'c3* = circle(p5(l2), p6(l2))',
    'c4* = circle(p6(l2), p5(l2))',
    'l3* = line(p7(c3, c4), p8(c3, c4))',
    'c5* = circle(p9(l3), p10(l3))',
    'c6* = circle(p10(l3), p9(l3))',
    'l4* = line(p11(c5, c6), p12(c5, c6))',
    'l5 = line(p13(l1, l2), p14(l2, l3))',
    'l6 = line(p14(l2, l3), p15(l3, l4))',
    'l7 = line(p15(l3, l4), p16(l1, l4))',
    'l8 = line(p16(l1, l4), p13(l1, l2))',
    
    eq_t
    'l1 = line(p1(), p2())',
    'c1* = circle(p1(), p2())',
    'c2* = circle(p2(), p1())',
    'l2 = line(p1(), p3(c1, c2))',
    'l3 = line(p2(), p3())',
    
    rhombus
    'l1* = line(p1(), p2())',
    'c1* = circle(p1(), p2())',
    'c2* = circle(p2(), p1())',
    'l2* = line(p3(c1, c2), p4(c1, c2))',
    'l3 = line(p1(), p3(c1, c2))',
    'l4 = line(p3(c1, c2), p2())',
    'l5 = line(p2(), p4(c1, c2))',
    'l6 = line(p4(c1, c2), p1())',
    parallel_l
    'l1* = line(p1(), p2())',
    'c1* = circle(p1(), p3(l1))',
    'c2* = circle(p3(l1), p1())',
    'c3* = circle(p2(), p3(l1))',
    'l2 = line(p4(c1, c2), p5(c1, c2))',
    'l3 = line(p6(c2, c3), p7(c2, c3))',
    
    rhomboid
    'l1* = line(p1(), p2())',
    'c1* = circle(p1(), p3(l1))',
    'c2* = circle(p3(l1), p1())',
    'c3* = circle(p4(l1, c2), p3(l1))',
    'c4* = circle(p5(l1, c3), p6(l1, c2))',
    'l2 = line(p1(), p7(c1, c2))',
    'l3 = line(p7(c1, c2), p8(c3, c4))',
    'l4 = line(p8(c3, c4), p4(l1, c2))',
    'l5 = line(p4(l1, c2), p1())',
    
    angle
    'l1 = line(p1(), p2())',
    'l2 = line(p2(), p3())',
    
    square
    'l1* = line(p1(), p2())',
    'c1* = circle(p1(), p2())',
    'c2* = circle(p2(), p1())',
    'l2* = line(p3(c1, c2), p4(c1, c2))',
    'c3* = circle(p5(l1, l2), p1())',
    'l3 = line(p1(), p6(l2, c3))',
    'l4 = line(p6(l2, c3), p2())',
    'l5 = line(p2(), p7(l2, c3))',
    'l6 = line(p7(l2, c3), p1())',
    
    sixty_ang
    'l1 = line(p1(), p2())',
    'c1* = circle(p1(), p2())',
    'c2* = circle(p2(), p1())',
    'l2 = line(p1(), p3(c1, c2))',
    
    right_ang_t
    'l1* = line(p1(), p2())',
    'c1* = circle(p1(), p2())',
    'c2* = circle(p2(), p1())',
    'l2* = line(p3(c1, c2), p4(c1, c2))',
    'l3 = line(p5(l1, l2), p6(l2))',
    'l4 = line(p5(l1, l2), p7(l1))',
    'l5 = line(p6(l2), p7(l1))',
    
    radii
    'c1 = circle(p1(), p2())',
    'l1 = line(p1(), p3(c1))',
    'l2 = line(p1(), p4(c1))',
    'l3 = line(p1(), p5(c1))',
    
    diameter
    'l1* = line(p1(), p2())',
    'c1* = circle(p1(), p2())',
    'c2* = circle(p2(), p1())',
    'l2 = line(p3(c1, c2), p4(c1, c2))',
    'c3 = circle(p5(l1, l2), p3(c1, c2))',
    
    triangle****
    'l1 = line(p1(), p2())',
    'l2 = line(p2(), p3())',
    'l3 = line(p3(), p1())',
    
    quadrilateral****
    'l1 = line(p1(), p2())',
    'l2 = line(p2(), p3())',
    'l3 = line(p3(), p4())',
    'l4 = line(p4(), p1())',
    
    segment
    'c1 = circle(p1(), p2())',
    'l1 = line(p3(c1), p4(c1))',
    
    llcc****
    'l1 = line(p1(), p2())',
    'l2 = line(p3(), p4())',
    'c3 = circle(p5(l1), p6(l1, l2))',
    'c4 = circle(p7(), p8(l1, l2))',
    
    ccl
    'c1 = circle(p1(), p2())',
    'c2 = circle(p3(c1), p4())',
    'l3 = line(p5(c1), p6(c1, c2))',
    
    ccc****
    'c1 = circle(p1(), p2())',
    'c2 = circle(p3(c1), p4(c1))',
    'c3 = circle(p5(c1), p6(c1, c2))',
    
    clcl
    'c1 = circle(p1(), p2())',
    'l2 = line(p3(c1), p4(c1))',
    'c3 = circle(p5(), p6(c1, l2))',
    'l4 = line(p7(c1), p8(c3))',
    
    clt
    'c1 = circle(p1(), p2())',
    'l2 = line(p3(c1), p4())',
    'l3 = line(p5(c1, l2), p6(c1))',
    'l4 = line(p6(), p7(c1))',
    'l5 = line(p7(), p5())',
    
    llt
    'l1 = line(p1(), p2())',
    'l2 = line(p3(), p4())',
    'l3 = line(p5(l1, l2), p6())',
    'l4 = line(p6(), p7())',
    'l5 = line(p7(), p5())',
    
    clcc****
    'c1 = circle(p1(), p2())',
    'l2 = line(p3(c1), p4(c1))',
    'c3 = circle(p5(c1), p6(c1))',
    'c4 = circle(p7(c1), p8(c1))',
    
    lllc
    'l1 = line(p1(), p2())',
    'l2 = line(p3(l1), p4())',
    'l3 = line(p5(l2), p6())',
    'c4 = circle(p7(l1, l2), p8(l3))',
    
    cccc****
    'c1 = circle(p1(), p2())',
    'c2 = circle(p3(c1), p4(c1))',
    'c3 = circle(p5(c1, c2), p6(c1, c2))',
    'c4 = circle(p7(c2, c3), p8(c2, c3))',
    
    cccl
    'c1 = circle(p1(), p2())',
    'c2 = circle(p3(), p4(c1))',
    'c3 = circle(p5(), p6(c1, c2))',
    'l4 = line(p7(c1, c2), p8(c1, c2))',
    
    clll
    'c1 = circle(p1(), p2())',
    'l2 = line(p3(c1), p4())',
    'l3 = line(p5(l2), p6())',
    'l4 = line(p7(c1, l2), p8(l3))',
    
    llll
    'l1 = line(p1(), p2())',
    'l2 = line(p3(), p4())',
    'l3 = line(p5(l1, l2), p6())',
    'l4 = line(p7(l1, l2), p8())',
    
    tll
    'l1 = line(p1(), p2())',
    'l2 = line(p1(), p3())',
    'l3 = line(p2(), p3())',
    'l4 = line(p4(l1, l2), p5())',
    'l5 = line(p6(l1, l2), p7())',
    
    cll
    'c1 = circle(p1(), p2())',
    'l2 = line(p3(c1), p4(c1))',
    'l3 = line(p5(c1), p6(c1))',
    
    lll
    'l1 = line(p1(), p2())',
    'l2 = line(p3(l1), p4())',
    'l3 = line(p5(l1), p6())',
    
    llc
    'l1 = line(p1(), p2())',
    'l2 = line(p3(), p4(l1))',
    'c3 = circle(p5(l1), p6(l1, l2))',
    
    cct
    'c1 = circle(p1(), p2())',
    'c2 = circle(p3(), p4())',
    'l3 = line(p5(c1, c2), p6(c1, c2))',
    'l4 = line(p6(), p7(c2))',
    'l5 = line(p7(), p5())',
    
    tcl
    'l1 = line(p1(), p2())',
    'l2 = line(p1(), p3())',
    'l3 = line(p2(), p3())',
    'c4 = circle(p4(l1, l2), p5())',
    'l5 = line(p6(l1, l2), p7(l3))',
    
    lcc
    'l1 = line(p1(), p2())',
    'c2 = circle(p3(l1), p4(l1))',
    'c3 = circle(p5(l1, c2), p6(l1, c2))',
    
    tcc
    'l1 = line(p1(), p2())',
    'l2 = line(p1(), p3())',
    'l3 = line(p2(), p3())',
    'c4 = circle(p4(l1, l2), p5())',
    'c5 = circle(p6(l1, l2), p7(l3))',
\end{verbatim}
\newpage 
\section{Geoclidean Context-Free Grammar}
\label{app:geoclidean_cfg}
Context free grammar used to parse Geoclidean programs into parse trees. 
\begin{verbatim}

        S -> ConstructionSequence
        
        ConstructionSequence -> Construction | Construction ConstructionSequence
        
        Construction -> ObjectDef '=' Object
        
        ObjectDef -> VisibleObject | InvisibleObject
        VisibleObject -> OBJECT_NAME
        InvisibleObject -> OBJECT_NAME '*'
        
        Object -> Line | Circle
        
        Line -> 'line' '(' PointRef ',' PointRef ')'
        Circle -> 'circle' '(' PointRef ',' PointRef ')'
        
        PointRef -> SimplePoint | ConstrainedPoint
        SimplePoint -> Point EmptyArgs
        ConstrainedPoint -> Point ConstraintArgs
        
        Point -> POINT_NAME
        
        EmptyArgs -> '(' ')'
        ConstraintArgs -> '(' Constraints ')'
        
        Constraints -> SingleConstraint | DoubleConstraint
        SingleConstraint -> ObjectRef
        DoubleConstraint -> ObjectRef ',' ObjectRef
        
        ObjectRef -> OBJECT_NAME | OBJECT_NAME '*'
        
        OBJECT_NAME -> LINE_NAME | CIRCLE_NAME
        LINE_NAME -> 'l1' | 'l2' | 'l3' | 'l4' | 'l5' | 'l6' | 'l7' | 'l8' | 'l9'
        CIRCLE_NAME -> 'c1' | 'c2' | 'c3' | 'c4' | 'c5' | 'c6'
        
        POINT_NAME -> 'p1' | 'p2' | 'p3' | 'p4' | 'p5' | 'p6' | 'p7' | 'p8' ...| 'p16'
\end{verbatim}
\clearpage
\section{Consistency of Geometric Regularity Effect Across Different Subjects}

\begin{figure}[h]
    \centering  
    \includegraphics[width=0.3\linewidth]{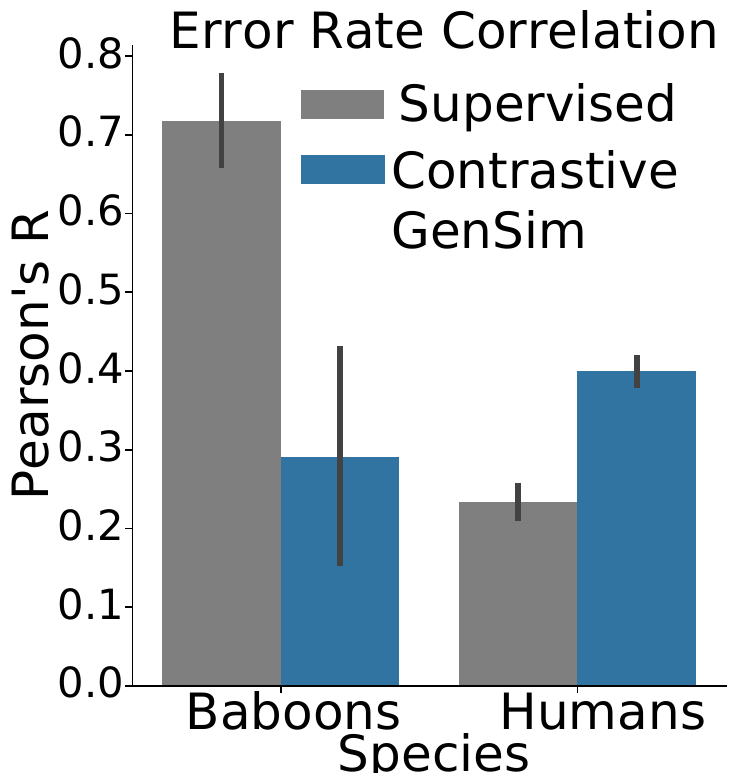}
  \caption{\textbf{Consistency of Figure~\ref{fig:quads}C results across different subjects} Correlation between mean finetuned models’ error rates with individual human or monkey error rates. This is a reproduction of Fig. 3C, but with behavior correlations for individual subjects instead of the average over subjects. Error bars denote 95\% confidence intervals over different subjects (different monkeys or different humans).  }
  \label{fig:consistency}
\end{figure}
\section{Accuracy of Natural Image Category Decoding Across WordNet Levels}
\begin{figure}[h]
    \centering  
    \includegraphics[width=0.6\linewidth]{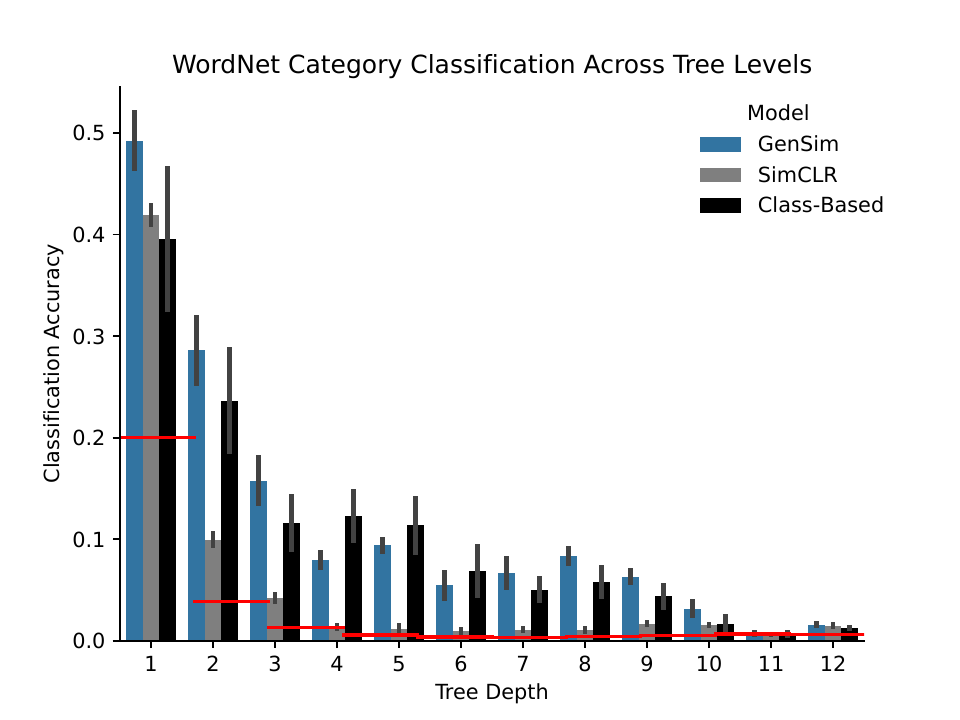}
  \caption{\textbf{Natural Image Category Classification Accuracy across all WordNet Tree Levels} All classification accuracies for the experiment in Figure~\ref{fig:imagenet}. Chance accuracies are denoted by red lines; note that these decrease as one goes deeper into the WordNet tree since the number of categories increases. Error bars denote standard error across model training runs.}
  \label{fig:tree_level_acc}
\end{figure}


\end{document}